\documentclass[numbered]{trbunofficial}
\usepackage[hidelinks]{hyperref}
\usepackage{graphicx}
\usepackage{tikz}
\usetikzlibrary{decorations.pathreplacing}
\usepackage{subcaption}
\usepackage{silence}
\usepackage{caption}
\usepackage{xcolor}
\usetikzlibrary{shapes.geometric, arrows, positioning}
\usepackage{float}

\tikzstyle{process} = [rectangle, rounded corners, minimum width=2.5cm, minimum height=1cm, text centered, draw=black, fill=gray!20]
\tikzstyle{decision} = [diamond, minimum width=2.5cm, minimum height=1cm, text centered, draw=black, fill=red!20]
\tikzstyle{arrow} = [thick,->,>=stealth]
\tikzstyle{startstop} = [rectangle, rounded corners, minimum width=2.5cm, minimum height=1cm, text centered, draw=black, fill=green!20]
\AuthorHeaders{Wang, Usama, Koutsopoulos, He}
\title{Estimating Link Level Traffic Emissions: Enhancing MOVES with Open-Source Data}

\author{%
  \textbf{Lijiao Wang}\\
  Department of Civil and Environmental Engineering\\
  Northeastern University, Boston, Massachusetts, 02115\\
  Email: wang.liji@northeastern.edu   \\
  \hfill\break%
  \textbf{Muhammad Usama, Ph.D.}\\
  Department of Civil and Environmental Engineering\\
  Northeastern University, Boston, Massachusetts, 02115\\
  Email: engrmusama90@gmail.com, m.usama@northeastern.edu \\
  \hfill\break
  \textbf{Haris N. Koutsopoulos, Ph.D.}\\
  Department of Civil and Environmental Engineering\\
  Northeastern University, Boston, Massachusetts, 02115\\
  Email: h.koutsopoulos@northeastern.edu\\
  \hfill\break%
  \textbf{Zhengbing He, Ph.D.}\\
  Laboratory for Information \& Decision Systems\\
  Massachusetts Institute of Technology, Cambridge, Massachusetts, 02139\\
  Email:  he.zb@hotmail.com, hezb@mit.edu
}

\begin{document}
\maketitle

\section{Abstract}

Open-source data offers a scalable and transparent foundation for estimating vehicle activity and emissions in urban regions. In this study, we propose a data-driven framework that integrates MOVES and open source GPS trajectory data, OpenStreetMap (OSM) road networks, regional traffic datasets and satellite imagery-derived feature vectors to estimate the link level operating mode distribution and traffic emissions. A neural network model is trained to predict the distribution of MOVES-defined operating modes using only features derived from readily available data. The proposed methodology was applied using open source data related to 45 municipalities in the Boston Metropolitan area. The ``{ground truth}'' operating mode distribution was established using OSM open source GPS trajectories. Compared to the MOVES baseline, the proposed model reduces RMSE by over 50\% for regional scale traffic emissions of key pollutants including CO, NO\textsubscript{x}, CO\textsubscript{2}, and PM\textsubscript{2.5}. This study demonstrates the feasibility of low-cost, replicable, and data-driven emissions estimation using fully open data sources.

\hfill\break%
\noindent\textit{Keywords}: {Traffic emissions, operating mode distribution, EPA MOVES, Modular neural networks, Open-source data, Data-driven modeling}
\newpage

\section{Introduction}

The transportation sector remains a dominant contributor to global energy consumption and greenhouse gas (GHG) emissions contributing 28\% of total U.S. GHG emissions in 2022 \cite{epa2025Fast}. The environmental, social, and economic ramifications of transportation-related emissions underscore the need for accurate and efficient traffic emissions estimation methods to inform policy and mitigation strategies.

Vehicle emissions are a complex phenomenon influenced by a range of factors, including complex driving behaviors, dynamic traffic conditions, the presence and type of traffic control, vehicle characteristics, fuel type, and prevailing ambient operating conditions \cite{wang2025estimating,zahoor2019LNG,Li2020ODevelopment,Zhang2012a}. Consequently, the development of emission models necessitates the incorporation of a diverse range of variables to accurately reflect the impact of these factors on emissions \cite{Ntziachristos2009COPERT,EPA2024MOVES}. Existing emission modeling approaches broadly fall into two categories: fuel-based and travel-based \cite{wang2025estimating}. Fuel-based models, such as the Computer Programme to Calculate Emissions from Road Transport (COPERT) developed by the European Environment Agency \cite{Ntziachristos2009COPERT}, directly leverage fuel consumption data to estimate GHGs based on emission factors per unit of fuel consumed. In contrast, travel-based emission models combine region-specific emission factors with travel activity data to generate comprehensive emission inventories. In the U.S., the primary models are the MOtor Vehicle Emission Simulator (MOVES) by the U.S. Environmental Protection Agency (EPA) \cite{EPA2024MOVES} and EMFAC by the California Air Resources Board (CARB) \cite{carb2025emfac}. 

The application of MOVES plays a crucial role in transportation and environmental planning. As a widely adopted tool, MOVES provides essential input for long-range transportation plans, air quality conformity assessments, and State Implementation Plans (SIPs) and transportation conformity analyses outside California \cite{Kondaru2018Generating}, enabling planners to evaluate the environmental impacts of future infrastructure investments, travel demand management strategies, and policy interventions. Its integration into planning workflows ensures that transportation decisions are aligned with national and regional air quality goals.

A critical component in emission models is the concept of driving cycles, also known as driving schedules. These are time-series data representing vehicle speed over time, designed to emulate real-world driving scenarios for assessing vehicle performance metrics like emissions and fuel consumption \cite{Tutuianu2015Development}. Driving cycles are broadly categorized as modal, with constant acceleration and speed phases, or transient, which feature frequent and dynamic speed changes \cite{wang2025estimating}. While standardized driving cycles exist, they often struggle to fully capture the rich diversity of real-world driving behaviors across varying geographies, traffic conditions, and driver behaviors. Significant research has focused on enhancing the accuracy and representativeness of driving cycles. In the U.S., the National Renewable Energy Laboratory (NREL) has contributed with tools like DRIVE \cite{NREL2024DRIVE}, which utilizes GPS data to generate custom driving cycles from real-world activity. Zhang et al. \cite{Zhang2022Heavy} analyzed the start and idling activities to refine MOVES emission estimates, emphasizing the need for fleet-specific data.

The MOVES model offers a range of approaches for estimating vehicle emissions at various scales, from national to project levels. It calculates emissions using, on one end, inputs like Vehicle Miles Traveled (VMT) for broader analyses, and on the other end, detailed second-by-second link-level speed profiles. The project-level approach, designed for link-level emissions, employs different pipelines depending on the data availability. For the highest accuracy, it uses second-by-second link average speed profiles derived from actual trajectory data to calculate operating mode distributions. However, such detailed speed profiles are often unavailable. In such cases, MOVES defaults to an alternative project-level approach, where aggregated link speeds are used to select predefined drive cycles as a proxy for estimating operating mode distributions. Since these default drive cycles may not reflect real-world operating conditions, they can lead to inaccurate emissions estimates—particularly for critical pollutants, e.g., PM$_{2.5}$ \cite{wang2025estimating}. This study focuses on the MOVES project-level approach by developing a neural network-based model that replaces the operating mode distribution estimation step of the MOVES. An important characteristic of the proposed model is that it leverages readily available open-source data to rapidly and accurately infer operating mode distributions across all network links, improving upon the default project-level MOVES approach.

Recent advances in machine learning have led to new approaches for emission estimation. Zahoor et al.~\cite{zahoor2019LNG} demonstrated that artificial neural networks could effectively predict emissions from LNG buses using vehicle operation data. Similarly, He et al.~\cite{He2020Estimating} developed a spatiotemporal cell-based model for CO\textsubscript{2} estimation on freeways using high-resolution traffic data. Howlader et al. \cite{HOWLADER2023Data} developed an integrated deep neural network method for predicting instantaneous vehicle emissions from light-duty vehicles, which are however limited to individual trajectories. Another study \cite{alam2025Deep} proposed a deep learning approach, enhanced by explainable artificial intelligence methods, to predict engine level CO\textsubscript{2} emissions based on various vehicle attributes and fuel consumption. These approaches show strong potential but often depend on proprietary data sources or simulated environments, which can be costly and difficult to replicate.


Prior research has also explored estimating operating mode distributions directly from macroscopic variables. Li et al. (2020) used a method to estimate operating mode distributions using average speed as input, demonstrating that even at similar average speeds, arterials and collectors exhibit distinct operating mode distributions \cite{Li2020ODevelopment}. However, their model was limited to average speed and did not incorporate the wealth of infrastructure-related features that influence real-world driving patterns.
A recent study proposed a Neural Networks-based framework to estimate the operating mode distribution of light-duty vehicles based on roadway infrastructure and traffic characteristics \cite{wang2025estimating}. The study demonstrated strong performance in predicting operating modes and associated emissions, with training data generated from detailed vehicle trajectories simulated using a microscopic traffic model. While simulation-based trajectories offer high fidelity, building city-wide simulation models is time-consuming, labor-intensive, and often infeasible for broad or multi-city applications due to the extensive data and calibration requirements.

Building on this work, this paper proposes a novel methodology to estimate city-wide operating mode distributions for traffic links within a network. Unlike previous efforts that relied on simulated traffic data, which is difficult to obtain and calibrate for city-scale models, this study directly leverages large-scale, open-source trajectory datasets derived from OpenStreetMap (OSM) Public GPS traces to train the operating mode distribution estimation method. This shift from simulated to actual GPS trajectories, provides a more direct and realistic source of data for training models to learn operating mode distributions. In addition to fundamental link-level infrastructure features, this study incorporates road grade data and satellite imagery, respectively obtained from elevation and Google APIs, as key inputs to the model, comprising feature vectors. Satellite imagery provides a rich representation of the topological, urban environment, land use, and network layout characteristics of an area. The imagery data can capture subtle factors that contribute to traffic emissions.

The proposed approach utilizes a Neural Network to learn the complex relationships between these diverse input features and the resulting operating mode distributions. This framework offers a more efficient and potentially more accurate alternative to the conventional MOVES approach. The research has significant implications, enabling accurate emissions estimation to support both research and policy at various scales. The key contributions are: 



\begin{itemize}
    \item Proposes a neural network-based approach to estimate operating mode distributions for traffic links based on open source data. The model integrates macroscopic traffic variables, detailed link infrastructure features, road grade, and satellite imagery features. 
    \item Introduces a framework for city-wide traffic emissions estimation that balances computational efficiency with accuracy, effectively overcoming the limitations of existing approaches.
\end{itemize}

The remainder of this paper is organized as follows: The data processing and the proposed model framework in presented in the Approach section. The Application section presents a case study to demonstrate the methodology and evaluate its performance with data from the Greater Boston Metropolitan area. The Conclusions section summarizes the key findings and discusses future research directions.

\section{METHODOLOGY}

\subsection{Model}

\begin{figure}[htbp]
    \centering
    \includegraphics[width=0.8\textwidth]{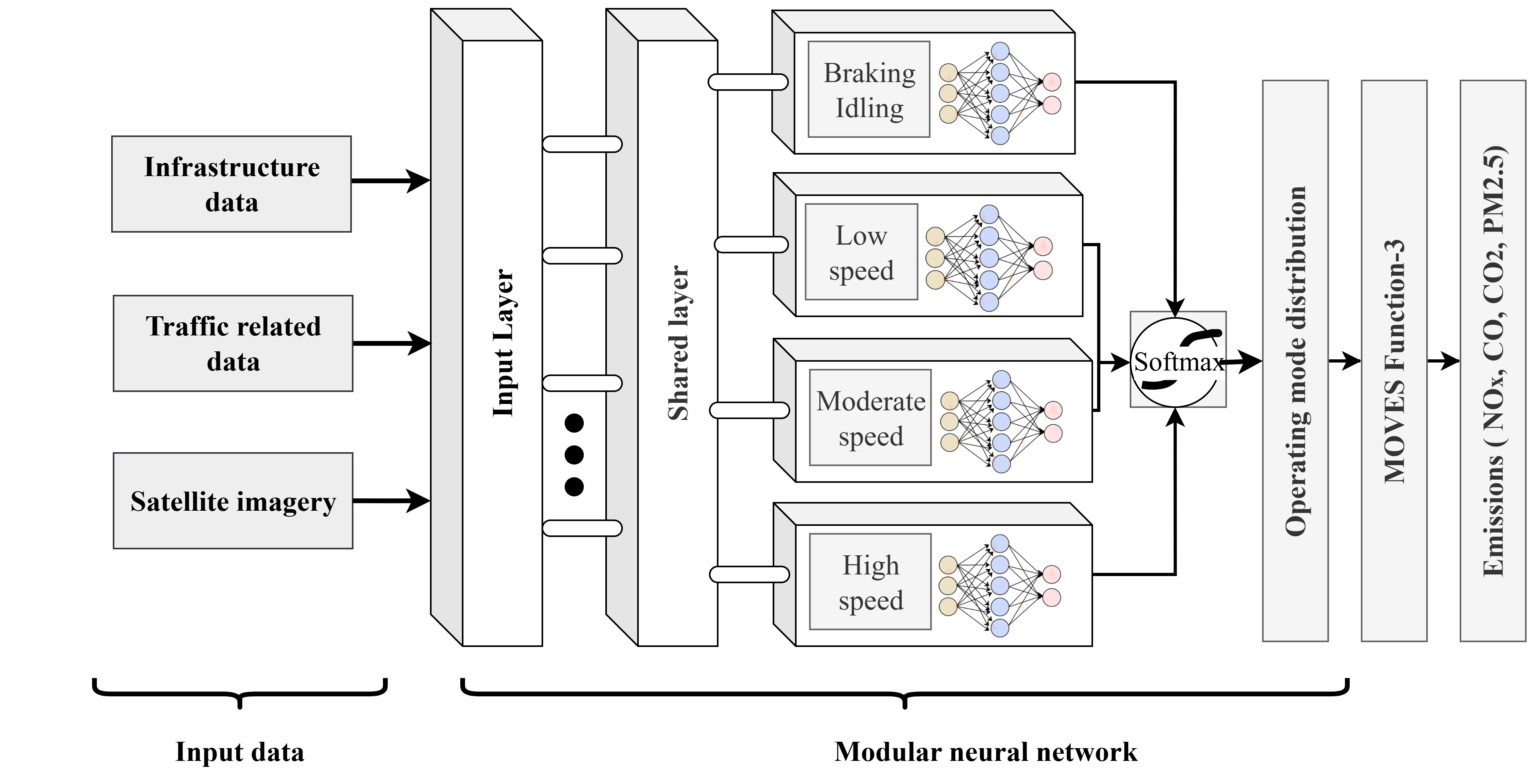}
    \caption{Methodological framework}
    \label{fig:framework}
\end{figure}

The comprehensive methodological framework for estimating link-level operating mode distributions, leveraging readily available open-source data, is presented in Figure~\ref{fig:framework}. The proposed model integrates infrastructure data, aggregated traffic, and satellite imagery to accurately predict operating mode distributions, which are empirically obtained from OSM Public GPS traces \cite{osmtraces}. 
The trajectory data serves exclusively for model training; once trained, the model can estimate operating mode distributions for any other city without requiring additional trajectory data. This estimated operating mode distribution is then used in conjunction with MOVES emission functions to derive pollutant emissions (e.g.,  NO\textsubscript{x}, CO, CO\textsubscript{2}, PM\textsubscript{2.5}), enabling more accurate link-level transportation emission assessments based on open data sources.

\begin{figure}[H]
    \centering
    \resizebox{\textwidth}{!}{
    \begin{tikzpicture}[xscale=-1]
    \begin{scope}[rotate=270] 
        \tikzstyle{unit}=[draw,shape=circle,fill=red!50,minimum size=30pt,inner sep=0pt]
        \tikzstyle{uniti} = [circle, draw, fill=blue!20, minimum size=30pt, inner sep=0pt]
        \tikzstyle{unith} = [circle, draw, fill=red!20, minimum size=30pt, inner sep=0pt]
        \tikzstyle{unitm} = [circle, draw, fill=orange, minimum size=30pt, inner sep=0pt]
        \tikzstyle{unito} = [circle, draw, fill=green, minimum size=30pt, inner sep=0pt]

        \def\xs{1.55} 
        \def\ys{1} 
 
        \node[uniti](x0) at (-1*\xs,3.5*\ys){\large$x_0$};
        \node[uniti](x1) at (-1*\xs,2*\ys){\large$x_1$};
        \node(dots) at (-1*\xs,1) {...};
        \node[uniti](xn) at (-1*\xs,0*\ys){\large$x_n$};
        \draw[draw=black, thick, rounded corners,fill=green!20] 
            (1.1,-1.25) rectangle (2.5,4.75); 
        \node[unith](h11) at (0.15*\xs,4.5){\large$h_1^1$};
        \node[unith](h12) at (0.15*\xs,2.5){\large$h_1^2$};
        \node(dots) at (0.15*\xs,1) {...};
        \node[unith](h1n) at (0.15*\xs,-1){\large$h_1^{128}$};

        \node[unith](h21) at (1.2*\xs,4){\large$h_2^1$};
        \node[unith](h22) at (1.2*\xs,2.5){\large$h_2^2$};
        \node(dots) at (1.2*\xs,1) {...};
        \node[unith](h2n) at (1.2*\xs,-0.5){\large$h_2^{64}$};

        \draw[draw=black, thick, rounded corners, fill=orange!20] 
            (2*\xs,8.25) rectangle (3.0*\xs,12.75);
        \node[unitm](sl1) at (2.5*\xs,12){\large$h_3^1$};
        \node[unitm](sl2) at (2.5*\xs,10.75){\large$h_3^2$};
        \node(dots) at (2.5*\xs,10){...};
        \node[unitm](sln) at (2.5*\xs,9){\large$h_3^{32}$};

        \node[unit](sl1_1) at (4*\xs,11.25){\large$h_7^1$};
        \node[unit](sl1_2) at (4*\xs,9.75){\large$h_7^2$};

        \draw[draw=black, thick, rounded corners, fill = blue!20] 
            (-0.75+2.5*\xs,2.25) rectangle (+0.75+2.5*\xs,6.75); 
        \node[unitm](sl21) at (2.5*\xs,6){\large$h_4^1$};
        \node[unitm](sl22) at (2.5*\xs,4.75){\large$h_4^2$};
        \node(dots) at (2.5*\xs,4){...};
        \node[unitm](sl2n) at (2.5*\xs,3){\large$h_4^{32}$};
        
        \draw[draw=black, thick, rounded corners,fill = yellow!40] 
            (-0.75+2.5*\xs,1.45) rectangle (+0.75+2.5*\xs,-2.9); 
        \node[unit](sl21_1) at (4*\xs,6){\large$h_8^1$};
        \node[unit](sl22_1) at (4*\xs,4.75){\large$h_8^2$};
        \node(dots) at (4*\xs,4){...};
        \node[unit](sl2n_1) at (4*\xs,3){\large$h_8^{6}$};

        \node[unitm](sl31) at (2.5*\xs,0.75){\large$h_5^1$};
        \node[unitm](sl32) at (2.5*\xs,-0.5){\large$h_5^2$};
        \node(dots) at (2.5*\xs,-1.25){...};
        \node[unitm](sl3n) at (2.5*\xs,-2.25){\large$h_5^{32}$};
        
        \draw[draw=black, thick, rounded corners,fill = purple!20] 
            (-0.75+2.5*\xs,-3.75) rectangle (0.75+2.5*\xs,-8); 
        
        \node[unit](sl31_1) at (4*\xs,0.75){\large$h_9^1$};
        \node[unit](sl32_1) at (4*\xs,-0.5){\large$h_9^2$};
        \node(dots) at (4*\xs,-1.25){...};
        \node[unit](sl3n_1) at (4*\xs,-2.05){\large$h_9^{9}$};
        
        \node[unitm](sl41) at (2.5*\xs,-4.5){\large$h_6^1$};
        \node[unitm](sl42) at (2.5*\xs,-5.75){\large$h_6^2$};
        \node(dots) at (2.5*\xs,-6.5){...};
        \node[unitm](sl4n) at (2.5*\xs,-7.25){\large$h_6^{32}$};

        \node[unit](sl41_1) at (4*\xs,-4.5){\large$h_{10}^1$};
        \node[unit](sl42_1) at (4*\xs,-5.75){\large$h_{10}^2$};
        \node(dots) at (4*\xs,-6.55){...};
        \node[unit](sl4n_1) at (4*\xs,-7.5){\large$h_{10}^{6}$};
        
        \node[unit, fill = orange!40](af) at (5.2*\xs,2){Softmax};

        \node[unito, scale=1.2](out1) at (6.5*\xs,7){\large$\hat{y}_{0}$};
        \node[unito, scale=1.2](out2) at (6.5*\xs,4.5){\large$\hat{y}_{1}$};
        \node[unito, scale=1.2](out3) at (6.5*\xs,2){\large$\hat{y}_{11}$};
          \node(dots) at (6.5*\xs,0){...};
        \node[unito, scale=1.2](out4) at (6.5*\xs,-2){\large$\hat{y}_{40}$};

        \draw[->] (af) -- (out1);
        \draw[->] (af) -- (out2);
        \draw[->] (af) -- (out3);
        \draw[->] (af) -- (out4);

        \draw[->] (sl1_2) -- (af);
        \draw[->] (sl1_1) -- (af);
        \draw[->] (sl22_1) -- (af);
        \draw[->] (sl21_1) -- (af);
        \draw[->] (sl2n_1) -- (af);
        \draw[->] (sl31_1) -- (af);
        \draw[->] (sl32_1) -- (af);
        \draw[->] (sl3n_1) -- (af);
        \draw[->] (sl41_1) -- (af);
        \draw[->] (sl42_1) -- (af);
        \draw[->] (sl4n_1) -- (af);

        \draw[->] (x0) -- (h11);
        \draw[->] (x0) -- (h12);
         \draw[->] (x0) -- (h1n);
 
        \draw[->] (x1) -- (h11);
        \draw[->] (x1) -- (h12);
         \draw[->] (x1) -- (h1n);
 
        \draw[->] (xn) -- (h11);
        \draw[->] (xn) -- (h12);
         \draw[->] (xn) -- (h1n);

         \draw[->] (h11) -- (h21);
        \draw[->]  (h11) -- (h22);
         \draw[->] (h11) -- (h2n);
         \draw[->] (h12) -- (h21);
        \draw[->]  (h12) -- (h22);
         \draw[->] (h12) -- (h2n);
         \draw[->] (h1n) -- (h21);
        \draw[->]  (h1n) -- (h22);
         \draw[->] (h1n) -- (h2n);

        \draw[->] (sl1) -- (sl1_1);
        \draw[->] (sl1) -- (sl1_2);

        \draw[->] (sl2) -- (sl1_1);
        \draw[->] (sl2) -- (sl1_2);

        \draw[->] (sln) -- (sl1_1);
        \draw[->] (sln) -- (sl1_2);

        \draw[->] (sl21) -- (sl21_1);
        \draw[->] (sl21) -- (sl22_1);
         \draw[->] (sl21) -- (sl2n_1);
 
        \draw[->] (sl22) -- (sl21_1);
        \draw[->] (sl22) -- (sl22_1);
         \draw[->] (sl22) -- (sl2n_1);
 
        \draw[->] (sl2n) -- (sl21_1);
        \draw[->] (sl2n) -- (sl22_1);
         \draw[->] (sl2n) -- (sl2n_1);

         \draw[->] (sl31) -- (sl31_1);
        \draw[->] (sl31) -- (sl32_1);
         \draw[->] (sl31) -- (sl3n_1);
 
        \draw[->] (sl32) -- (sl31_1);
        \draw[->] (sl32) -- (sl32_1);
         \draw[->] (sl32) -- (sl3n_1);
 
        \draw[->] (sl3n) -- (sl31_1);
        \draw[->] (sl3n) -- (sl32_1);
         \draw[->] (sl3n) -- (sl3n_1);

        \draw[->] (sl41) -- (sl41_1);
        \draw[->] (sl41) -- (sl42_1);
         \draw[->] (sl41) -- (sl4n_1);
 
        \draw[->] (sl42) -- (sl41_1);
        \draw[->] (sl42) -- (sl42_1);
         \draw[->] (sl42) -- (sl4n_1);
 
        \draw[->] (sl4n) -- (sl41_1);
        \draw[->] (sl4n) -- (sl42_1);
         \draw[->] (sl4n) -- (sl4n_1);
         

         \draw[->, thick] 
            (2.5,2.5) -- (2.95,10.5);
        \draw[->, thick] 
            (2.5,2.5) -- (2.955,4.5);
        \draw[->, thick] 
            (2.5,2.5) -- (2.95,-.75);
        \draw[->, thick] 
            (2.5,2.5) -- (2.95,-5.75);
 
        \draw [decorate,decoration={brace,amplitude=10pt,mirror},xshift=-4pt,yshift=0pt] (-2,11.5) -- (-1,11.5) node [black,midway,yshift=+0.cm, xshift = -2cm,rotate=0]{Input layer};

        \draw [decorate,decoration={brace,amplitude=10pt,mirror},xshift=-4pt,yshift=0pt] (0,11.5) -- (2.5,11.5) node [black,midway,yshift=+0.cm, xshift = -2cm,rotate=0]{Shared layers};
        \draw [decorate,decoration={brace,amplitude=10pt,mirror},xshift=-4pt,yshift=0pt] (3,13.0) -- (6,13.0) node [black,midway,xshift=-0.7cm,rotate=90]{Module-1};
        \draw [decorate,decoration={brace,amplitude=10pt,mirror},xshift=-4pt,yshift=0pt] (3,7) -- (6,7.0) node [black,midway,xshift=-0.7cm,rotate=90]{Module-2};
        \draw [decorate,decoration={brace,amplitude=10pt,mirror},xshift=-4pt,yshift=0pt] (3,1.45) -- (6,1.45) node [black,midway,xshift=-0.6cm,rotate=90]{Module-3};
        \draw [decorate,decoration={brace,amplitude=10pt,mirror},xshift=-4pt,yshift=0pt] (3,-3.75) -- (6,-3.75) node [black,midway,xshift=-0.6cm,rotate=90]{Module-4};
        \draw [decorate, decoration={brace, amplitude=10pt,mirror}]
        (9.0,11) -- (11,11)

    node [black, midway, yshift=0cm, align=center,  xshift = -2cm] {Output layer: \\ operating mode \\ distribution};

    \node[draw, rectangle, rounded corners, fill=white, align=center] at (1.8,-3) {Dropout\\($p$)};
    \node[draw, rectangle, rounded corners, fill=white, align=center] at (4,-9) {Dropout\\($p$)};
        
    \end{scope}
    \end{tikzpicture}
    }
    \caption{Modular neural network architecture with input nodes, hidden layers, and output layer with drop out ratio, adopted from \cite{wang2025estimating}.}
    \label{fig:model}
\end{figure}
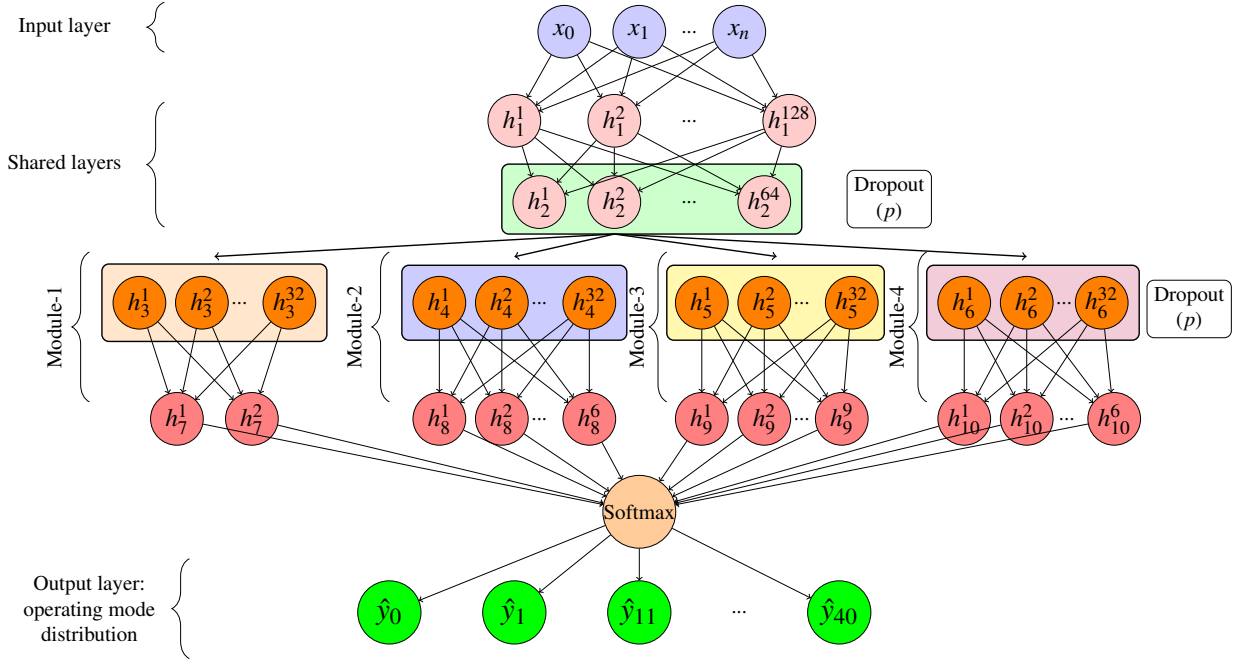

The proposed framework uses a modular neural network (MNN) architecture to estimate city-wide operating mode distributions of light-duty vehicles. As illustrated in Figure~\ref{fig:model}, the MNN consists of an input layer, two shared fully connected layers, four specialized modules corresponding to different speed groups, and an output layer with softmax normalization. Each module is responsible for estimating the probability distribution over a specific group of operating modes. The proposed model incorporates dropout regularization to improve generalizability and reduce the risk of overfitting due to increased data heterogeneity. 

The following sections detail the input datasets, their preprocessing, and the model architecture and training procedures.



%
\subsection{Data}
\subsubsection{Infrastructure}


Infrastructure data includes attributes such as road type, link length, directionality, speed limit, and number of lanes. All these features are retrieved from OSM. Raw OSM data does not provide predefined link definitions suitable for traffic modeling. While OSM organizes roadways into “ways” as sequences of connected nodes identified by a unique ID, these segments are often misaligned with the needs of traffic analysis. Some OSM ways span multiple intersections, creating overly long links that obscure fine-scale traffic dynamics, while others represent short segments that fail to capture meaningful roadway portions.

To address these limitations, we construct the road network using NetworkX~\cite{networkx_website}, a Python library for network analysis. NetworkX builds a graph-based representation of the OSM network, where nodes represent intersections or endpoints, and edges represent individual road segments between them. This results in a more granular and topologically consistent definition of links, often aligning with the intersection-to-intersection geometry of real-world roads. Compared to raw OSM ways, these links offer a more practical unit for traffic analysis, emissions modeling, and trajectory mapping, as they better reflect functional changes in the network and enable integration with diverse transportation datasets.

Road grade is calculated for each link edge by extracting the latitude and longitude of the edge’s start and end nodes. The Open-Elevation API~\cite{openelevationapi} is used to obtain the altitude at both nodes. The grade is computed as the ratio of the elevation difference to the horizontal distance between the two nodes, representing the slope of the edge.

\subsubsection{Traffic}

Aggregated traffic data, including Average Annual Daily Traffic (AADT) and average speed, is available at both national and state levels. Nationwide data comes from the Highway Performance Monitoring System \cite{fhwahpms}, while state-level data is provided through state traffic inventories and state planning models. Since these spatial units do not always align with the processed OSM network, we employ spatial joins supplemented by street name similarity analysis to ensure accurate dataset integration and mismatch resolution.

The peak hour flow (for peak hour emissions analysis) is estimated from the AADT using the following equation adopted from Highway Capacity Manual~\cite{highway_capacity_manual}:

\begin{linenomath}
\begin{equation}
\quad \text{Peak Hour Flow} = {AADT} \times K \times D
\end{equation}
\end{linenomath} \\
where \( K \) is the proportion of daily traffic occurring during the peak hour, and \( D \) is the directional distribution factor.

Subsequently, the Bureau of Public Roads (BPR) function is used to estimate corresponding link speeds using the peak hour flow ~\cite{bpr1964}. The BPR function relates link travel time to the volume and capacity of the road segment and is given by:

\begin{linenomath}
\begin{equation}
\quad t_a = t_a^0 \left( 1 + \alpha \left( \frac{v_a}{c_a} \right)^\beta \right)
\end{equation}
\end{linenomath}\\
where \( t_a \) is the actual travel time on link \( a \), \( t_a^0 \) is the free-flow travel time \( a \), \( v_a \) is the traffic volume on link \( a \), \( c_a \) is the capacity of link \( a \), and \( \alpha \) and \( \beta \) are calibration parameters, typically set to \( \alpha = 0.15 \) and \( \beta = 4 \). 


\subsubsection{Satellite Imagery}
Satellite imagery can provide important information on factors that impact emissions and complements other variables. To better capture the spatial characteristics of each study area, we incorporate high-resolution satellite imagery to represent the physical layout of each town. These images were retrieved using the \textit{Google Maps Static API}~\cite{google_api}, which provides consistent and detailed aerial views across geographic regions. To delineate the town boundaries accurately, we used the \textit{OSMnx} Python package~\cite{boeing2017osmnx} to download administrative boundary data from OSM. The satellite images were then clipped using the town boundaries to generate focused, town-specific layout maps.



\begin{figure}[htbp]
    \centering
    \includegraphics[width=0.9\textwidth]{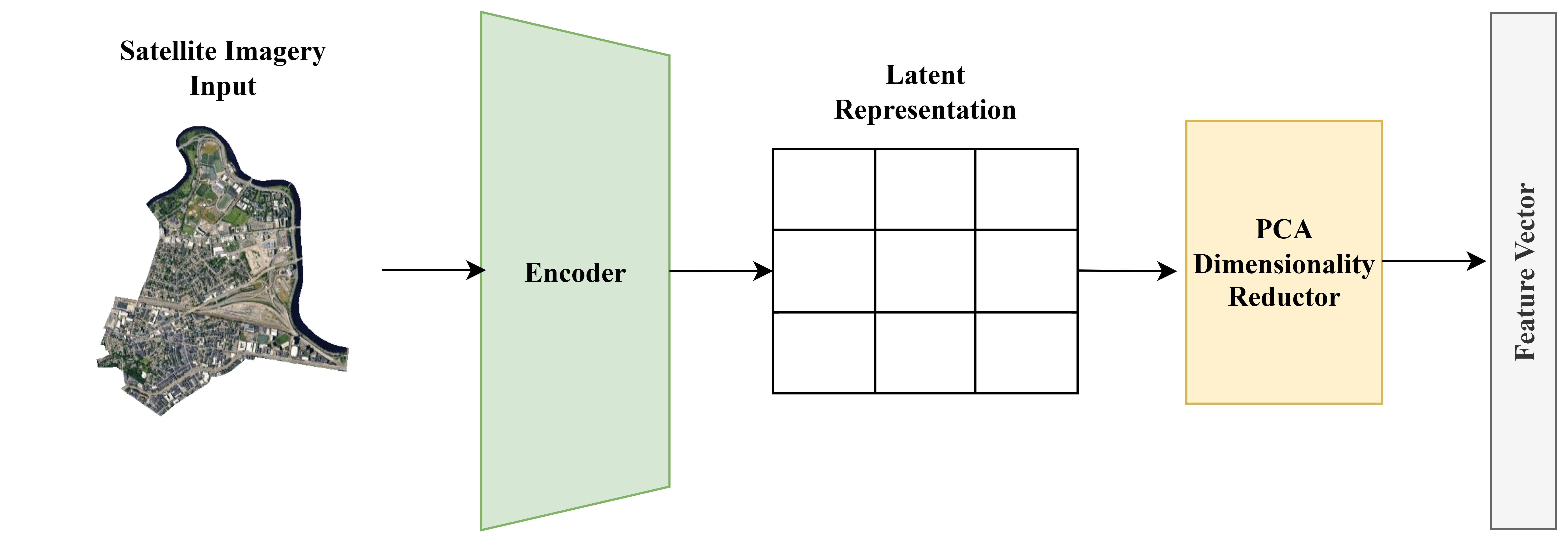}
    \caption{A framework that encodes satellite imagery into feature vector.}
    \label{fig:autoencoder_framework}
\end{figure}

To further leverage the spatial richness of these satellite images, we extracted numerical representations through a computer vision pipeline. Figure~\ref{fig:autoencoder_framework} illustrates the computer vision pipeline that is used to process the satellite images. The encoder transforms satellite imagery into latent features, and the decoder reconstructs key spatial indicators in vector form. This architecture enables the model to capture information related to road structure, land use patterns, and built environment features. Specifically, we used a pretrained ResNet-18 convolutional neural network ~\cite{paszke2019pytorch} to generate feature vector representations. To reduce dimensionality and eliminate redundant information, we applied Principal Component Analysis (PCA)~\cite{jolliffe2002principal} to the extracted features. We selected the number of principal components required to preserve 95\% of the cumulative explained variance, resulting in compact representations. This approach enabled us to visualize key elements of the built environment—such as road network structure, land use density, and overall urban form—that are not readily captured in numerical datasets. These spatial layouts provide context for interpreting variations in GPS trajectories, traffic flow, and emissions. For example, towns with compact, grid-like street networks may experience lower vehicle speeds and more frequent stops, whereas suburban layouts may lead to longer and smoother trips.

Table~\ref{tab:infra_variables} summarizes all model input features derived from open-source data, including variable definitions, data sources, and processing methods. These features, which span infrastructure attributes, traffic characteristics, and satellite imagery vectors, form the basis for the input data set used in this study, to train the operating mode distribution model.

\begin{table}[htbp]
\centering
\caption{Data and associate open sources}
\begin{tabular}{|l|p{6cm}|p{5cm}|}
\hline
\textbf{Variable} & \textbf{Definition} & \textbf{Data Source} \\
\hline
Road Type & OSM road classification & OSM \\
\hline
Segment Length & Length of each OSM edge (meters) & OSM \\
\hline
One-Way  & Whether the road allows traffic in one direction only & OSM \\
\hline
Speed Limit & Posted speed limit & OSM \\
\hline
Number of Lanes & Number of lanes on the roadway segment & OSM \\
\hline
Urban Type & Classification of area (urban, suburban, rural) & State DOT \\
\hline
Federal Functional Class & Federal road classification code & State DOT \\
\hline
AADT & Annual Average Daily Traffic & State DOT \\
\hline
VMT & Vehicle Miles Traveled & State DOT \\
\hline
Free-flow Speed  & Travel time under uncongested conditions & State DOT \\
\hline
Capacity  & Maximum hourly flow the road can handle & State DOT \\
\hline
Peak Hour Flow & \( \text{AADT} \times K \times D \), traffic in peak hour & Computed using AADT, \(K\), and \(D\)  \\
\hline
Grade & Calculated from elevation and horizontal distance & OSM + Elevation API \\
\hline
Travel Time & Estimated from BPR function & BPR function \\
\hline
Speed & Road length divided by \( t_a \) & Computed \\
\hline

Satellite Imagery Vector & Vector representation of town environment features extracted from satellite images using ResNet-18 and reduced via PCA & Google Maps Static API + OSMnx + PyTorch + scikit-learn \\
\hline

\end{tabular}
\label{tab:infra_variables}
\end{table}

\subsection{Training}


While local trajectory data can provide accurate operating mode distributions, to train the proposed NN model, ground-truth data, in terms of the distribution of the operating mode, is required. For this purpose, the detailed MOVES procedure is used. However, this requires detailed trajectory data along the link of interest. To address this, we leverage open-source trajectory data to establish a ground truth for operating mode distributions. The proposed neural network model is trained against this ground truth, using the readily available infrastructure, aggregated traffic, and satellite imagery as inputs.


\subsubsection{Trajectory Data}
The OSM public GPS trajectories~\cite{osmtraces} are valuable data sets that capture the movement of vehicles over time using latitude and longitude points. However, these raw trajectories often contain errors and inconsistencies, such as GPS drift, missing values, and inaccurate positioning. Additionally, the devices used to report OSM trajectories are not unified, leading to variations in temporal resolution and data quality.

    

\begin{figure}[htbp]
    \centering
    \includegraphics[width=0.8\textwidth]{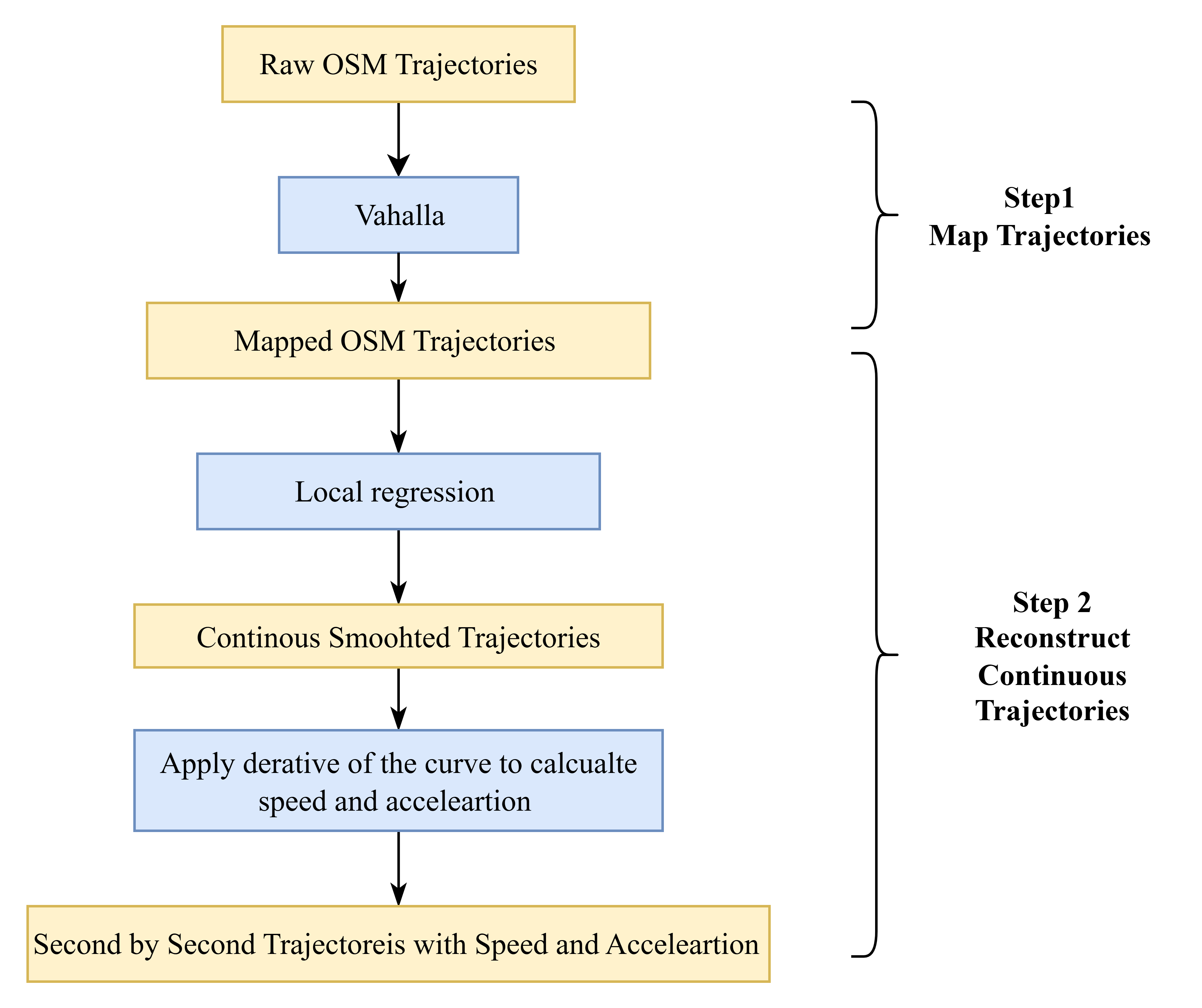}
    \caption{Workflow for preprocessing OSM GPS trajectories, including map matching using Valhalla and smoothing with local regression.}
    \label{fig:preprocessing_workflow}
\end{figure}

To address these spatial and temporal inaccuracies, we apply a two-step preprocessing approach, as illustrated in Figure~\ref{fig:preprocessing_workflow}. To ensure spatial accuracy and reduce noise in raw GPS data, we first perform map matching using the Valhalla open-source routing engine~\cite{valhalla}, which supports multi-modal navigation and OSM-based map alignment. This process projects each GPS point from the trajectory onto the most plausible segment of the OSM road network, effectively correcting for common location errors such as GPS drift or off-road positioning. By aligning the GPS traces with actual road geometry, this step improves the reliability of downstream feature extraction and trajectory-based modeling.

\begin{figure}[htbp]
    \centering
    \includegraphics[width=0.9\linewidth]{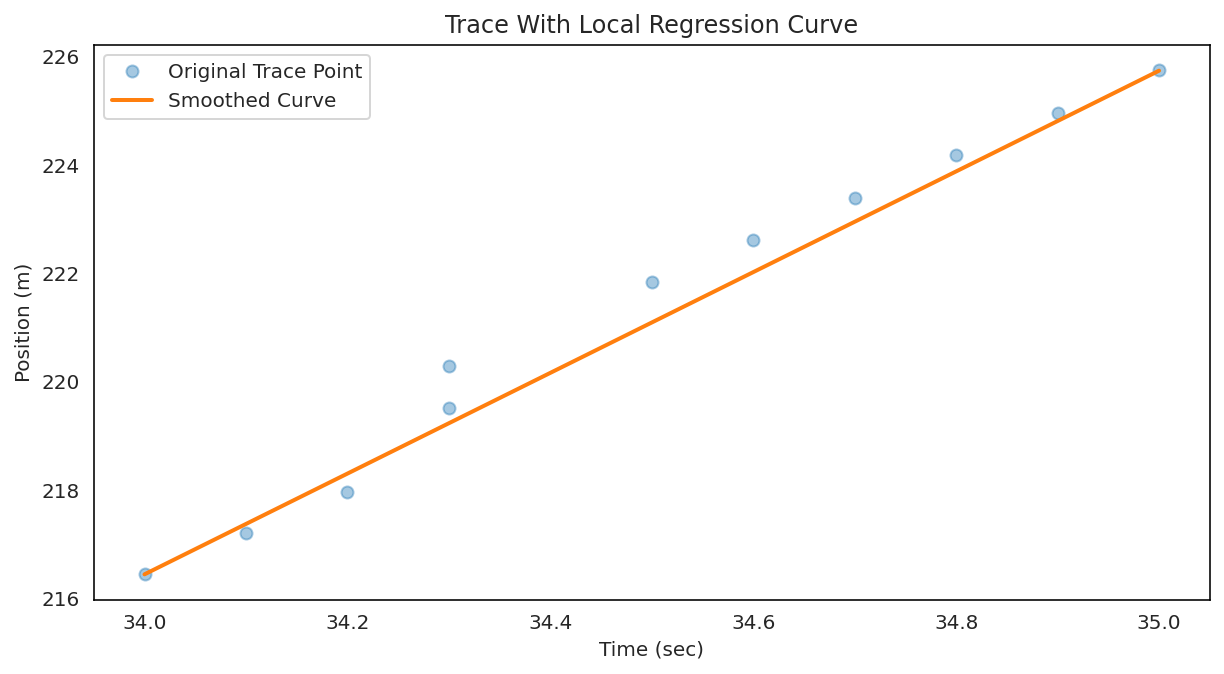}
    \caption{Raw GPS Positions and Smoothed Curve in One Second}
    \label{fig:trajectory_smoothing}
\end{figure}

Following the map matching step, we apply local regression smoothing to address discontinuities and noise in the GPS trajectories. This approach, inspired by the method proposed in~\cite{toledo2007estimation}, transforms raw and often jagged GPS point sequences into continuous and realistic representations of vehicle movement. As illustrated in Figure~\ref{fig:trajectory_smoothing}, OSM-based GPS trajectories are typically recorded with non-uniform and high-frequency time resolution. When GPS measurements contain bias, this high sampling rate can amplify noise in derived dynamic features such as speed and acceleration. The local regression to fit smooth curves that better reflect true motion addresses this issue effectively. Long temporal gaps, caused by data loss or extended stops, are handled by segmenting the trajectory to avoid unrealistic interpolation. Shorter interruptions, such as brief GPS signal loss or stop-and-go behavior at intersections, are smoothed by fitting localized curves. This process improves both spatial and temporal coherence, ensuring that                                                   the resulting trajectories are suitable for downstream analysis and feature extraction.

Once the trajectories are smoothed, we compute second-by-second speed and acceleration by taking the time derivatives of position. This enables consistent and detailed characterization of vehicle motion across devices with varying sampling rates. The resulting speed, acceleration, and road grade data are then used to assign operating mode distributions as defined in the MOVES model. These distributions serve as training targets for the proposed operating mode estimation framework.

\section{Application}

\begin{figure}[htbp]
    \centering
    \includegraphics[width=0.75\textwidth]{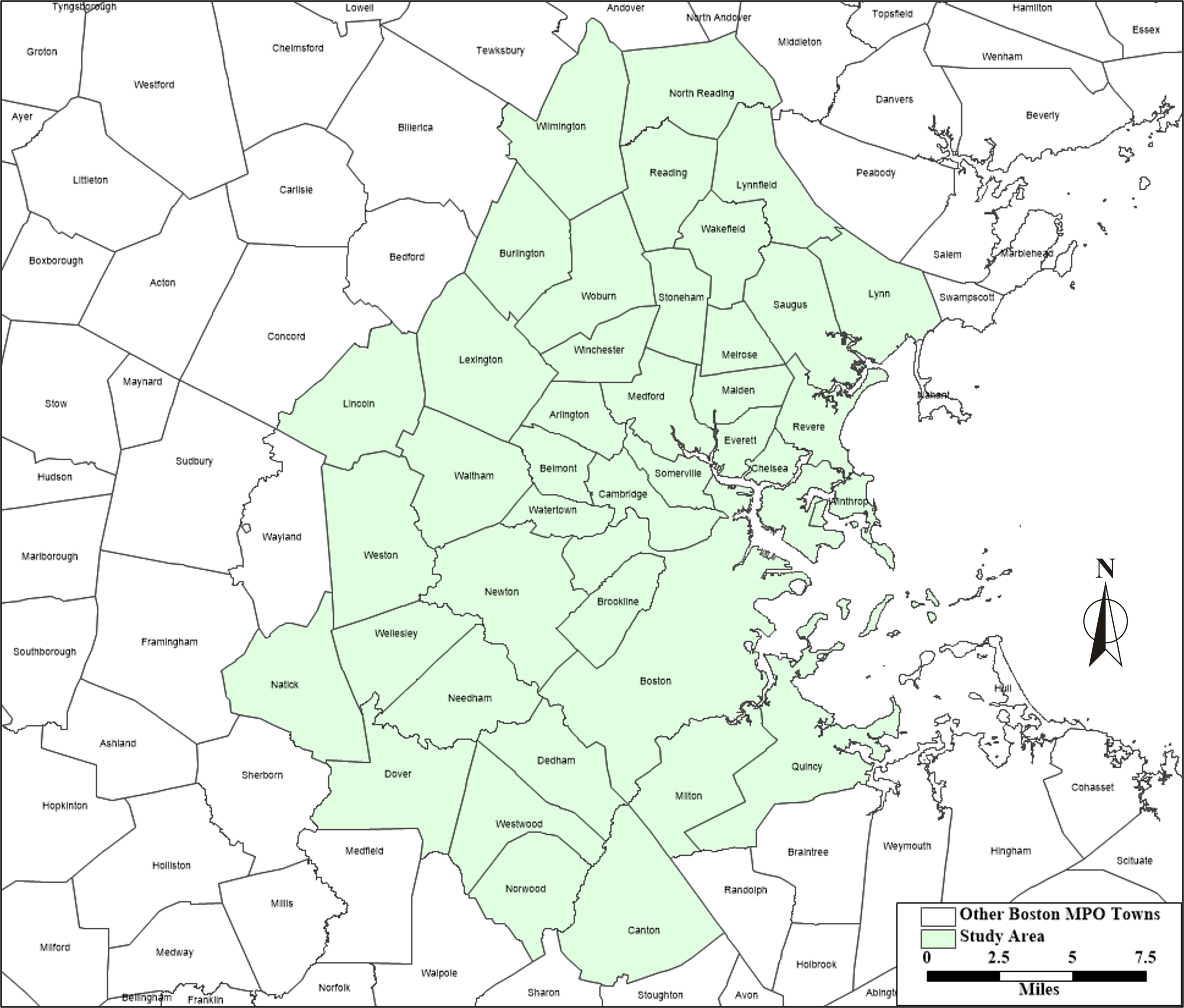}
    \caption{Study Area}
    \label{fig:MPO}
\end{figure}

For the training and evaluation of our proposed model, we utilized a study area comprising 45 towns situated within the Boston Metropolitan Area, shown in Figure \ref{fig:MPO} with green shade. This region offers a rich dataset for model application due to its significant spatial heterogeneity. It spans the full spectrum of urban development, from the high-density traffic and complex infrastructure of urban cores (e.g., downtown Boston) to the more diffuse patterns of suburban and rural settings. This diverse range of traffic and infrastructure characteristics inherent to the selected towns is critical for demonstrating the model's adaptability and its accurate performance across varying scales of urban development.

\subsection{Data preprocessing}

The raw OSM network is preprocessed to standardize link definitions before mapping infrastructure and traffic features. Specifically, NetworkX is employed to convert the raw data into a graph where each edge represents a link connecting two intersections.

\begin{figure}[htbp]
  \centering
  \begin{subfigure}[t]{0.48\textwidth}
    \centering
    \includegraphics[width=\linewidth]{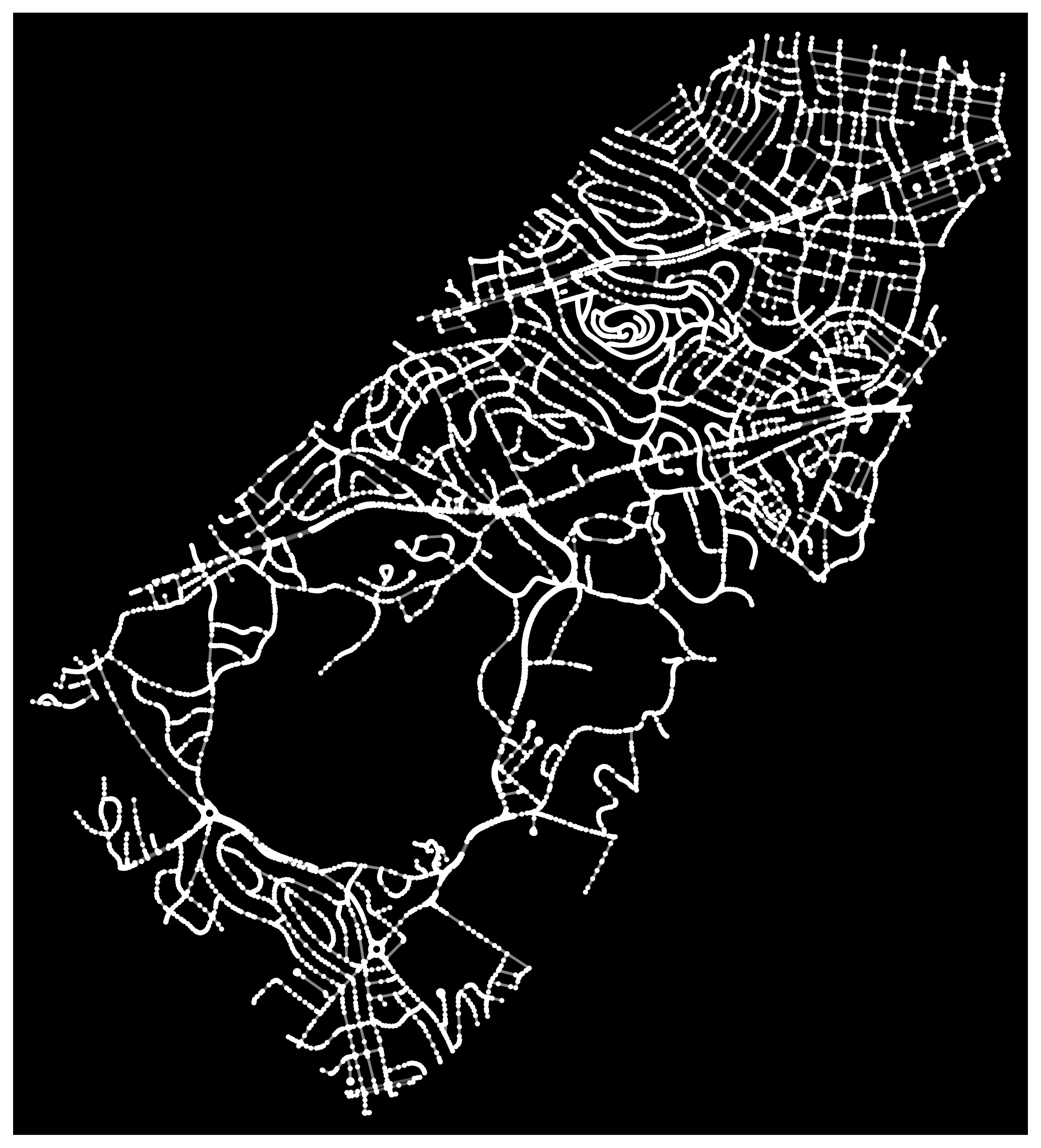}
    \caption{Raw OSM road network}
    \label{fig:raw_osm}
  \end{subfigure}
  \hfill
  \begin{subfigure}[t]{0.48\textwidth}
    \centering
    \includegraphics[width=\linewidth]{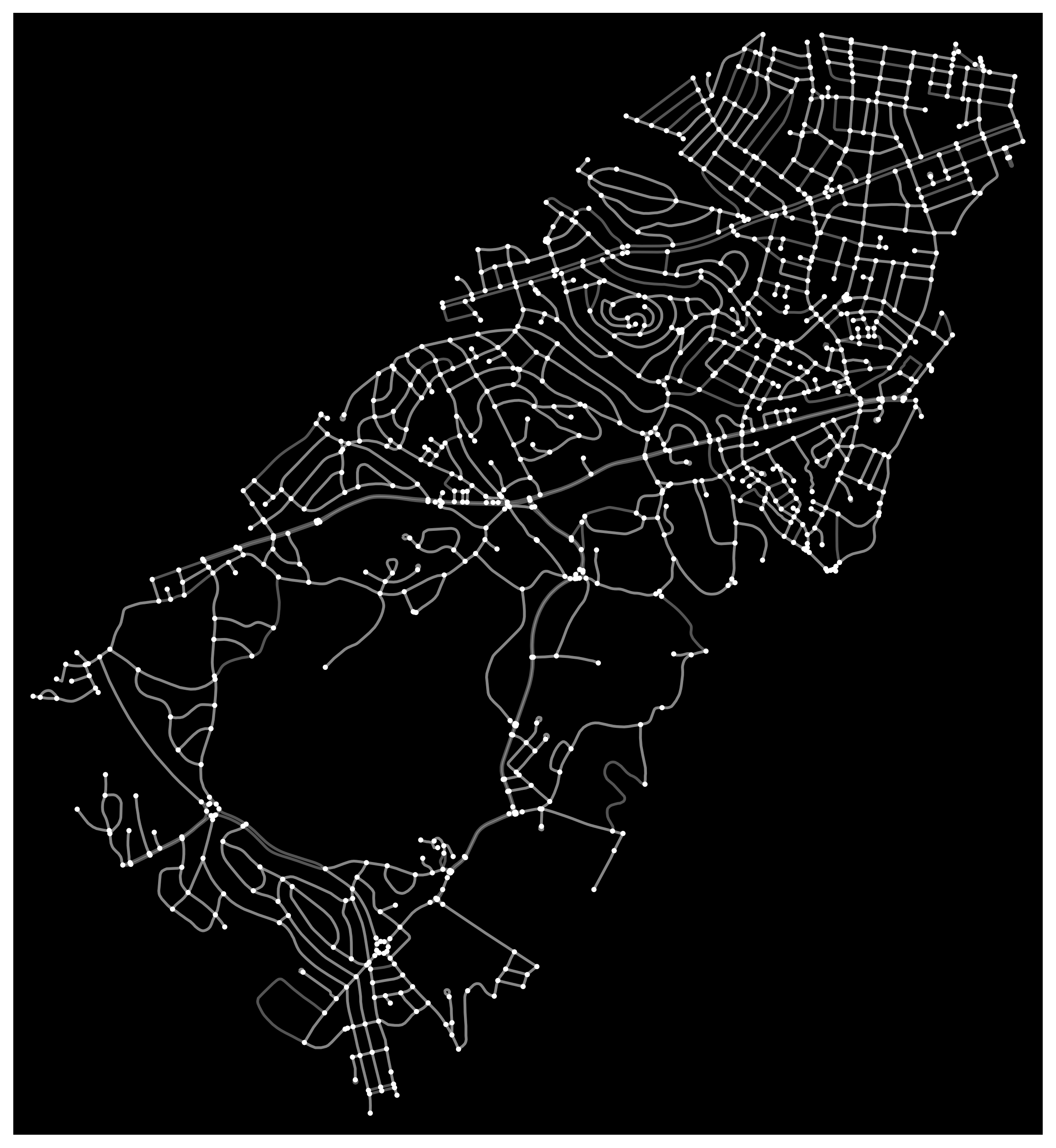}
    \caption{Processed road network}
    \label{fig:networkx_osm}
  \end{subfigure}
  \caption{Comparison of raw and processed road network representations for Brookline, MA.}
  \label{fig:network_comparison}
\end{figure}

Figure~\ref{fig:network_comparison} compares the raw OSM representation in Figure \ref{fig:raw_osm} with the processed network in Figure \ref{fig:networkx_osm} for Brookline. While raw OSM ``{ways}'' often span multiple intersections or appear as fragmented segments, processing enforces consistent link definitions—each edge connects precisely two nodes i.e., intersections or endpoints.

Following network processing, a comprehensive set of infrastructure and traffic features is extracted for each link from OSM and MassDOT data sources. The infrastructure attributes collected at the link level include: road type, length, road grade, number of lanes, directionality (one-way or two-way), posted speed limit, capacity and urban type. For traffic characteristics, we incorporate congested speeds and AADT for each link.

\begin{figure}[h]
    \centering
    
    \begin{subfigure}[b]{0.30\textwidth}
        \includegraphics[width=\textwidth]{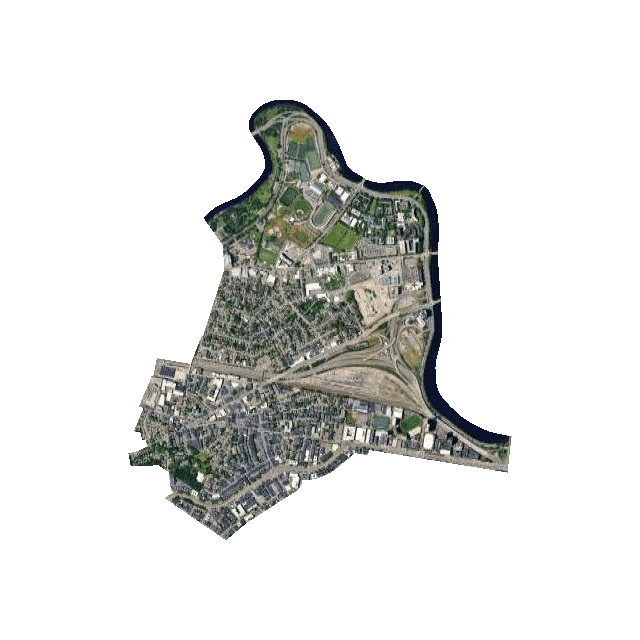}
        \caption{Allston}
    \end{subfigure}
    \hspace{0.01\textwidth}
    \begin{subfigure}[b]{0.30\textwidth}
        \includegraphics[width=\textwidth]{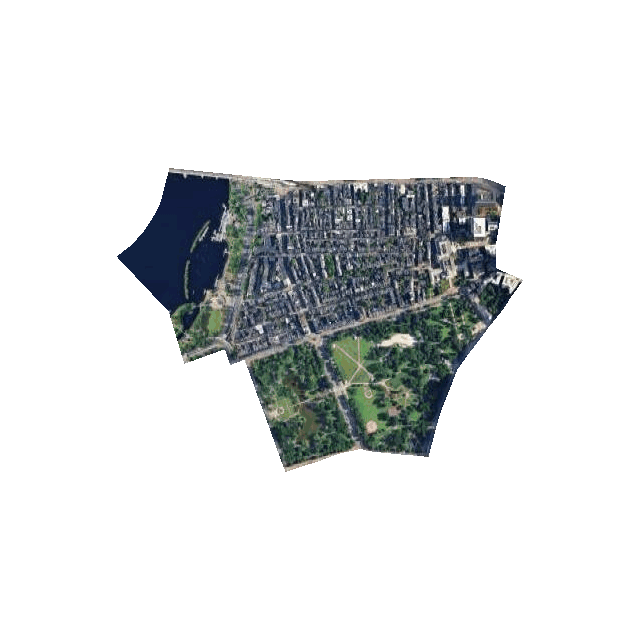}
        \caption{Beacon Hill}
    \end{subfigure}
    \hspace{0.01\textwidth}
    \begin{subfigure}[b]{0.30\textwidth}
        \includegraphics[width=\textwidth]{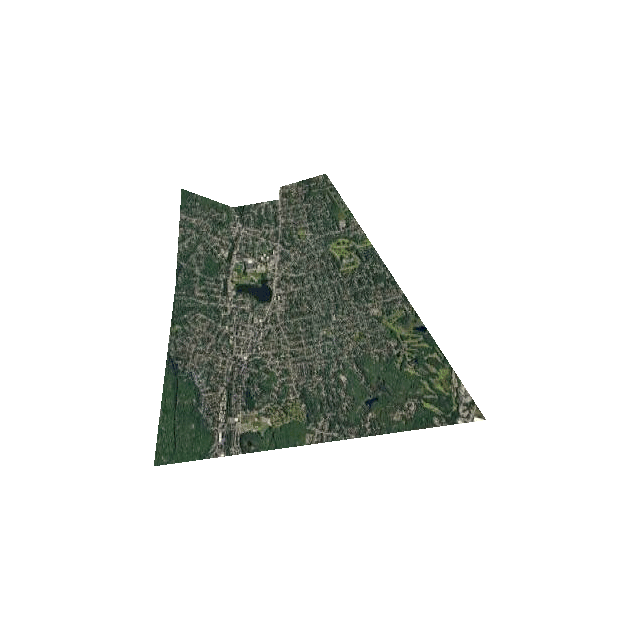}
        \caption{Melrose}
    \end{subfigure}
    
    \vspace{0.2cm}
    
    \begin{subfigure}[b]{0.30\textwidth}
        \includegraphics[width=\textwidth]{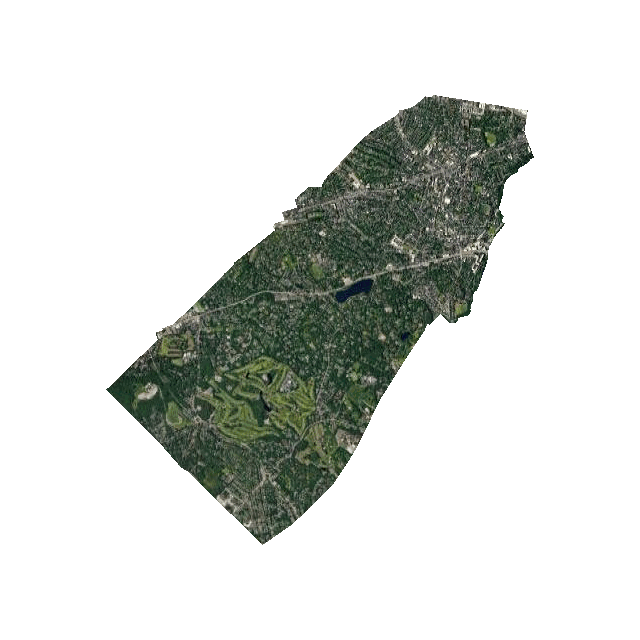}
        \caption{Brookline}
    \end{subfigure}
    \hspace{0.01\textwidth}
    \begin{subfigure}[b]{0.30\textwidth}
        \includegraphics[width=\textwidth]{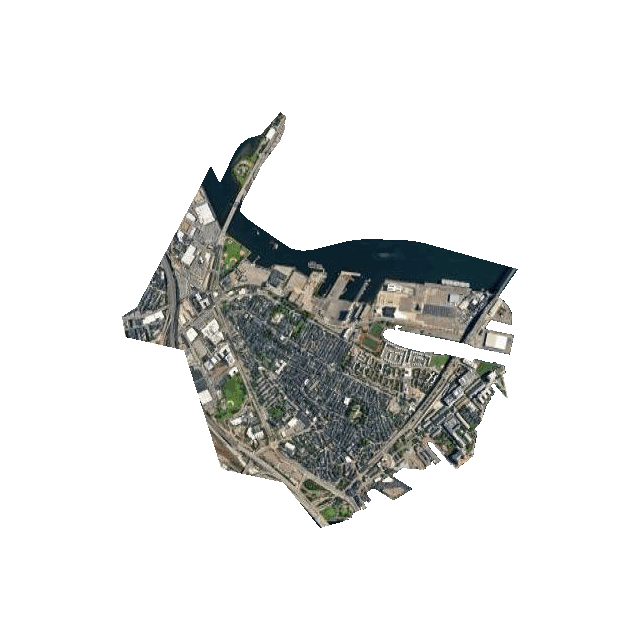}
        \caption{Charlestown}
    \end{subfigure}
    \hspace{0.01\textwidth}
    \begin{subfigure}[b]{0.30\textwidth}
        \includegraphics[width=\textwidth]{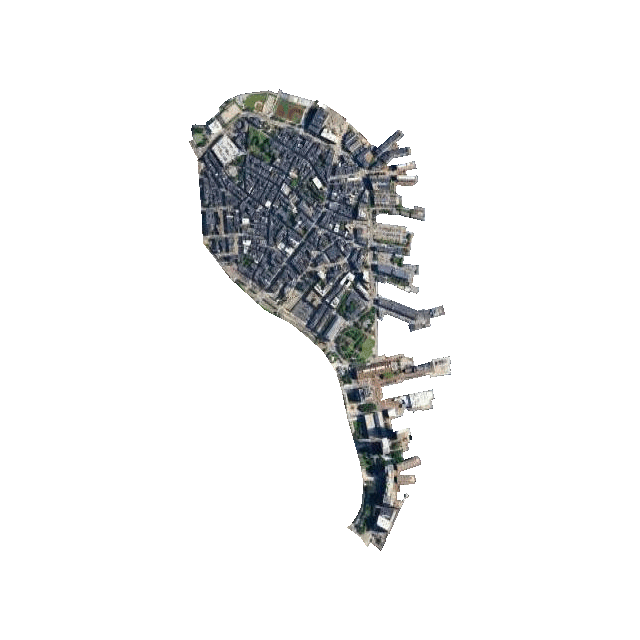}
        \caption{North End}
    \end{subfigure}
    
    \caption{Examples of town layout differences across study areas. Each image shows the built environment and road network structure extracted from high-resolution satellite imagery.}
    \label{fig:town_layout_examples}
\end{figure}

The satellite imagery is processed to get town level urban form features. Figure~\ref{fig:town_layout_examples} shows the variation in urban form across six representative towns. The imagery highlights differences in road network structure, block size, and land use patterns. Allston and North End show compact, grid-like layouts with dense development and minimal open space—characteristics typical of older urban cores. In contrast, Melrose and Brookline exhibit more dispersed suburban patterns, with winding roads, larger parcels, and greater green space. Beacon Hill and Charlestown reflect historically evolved street networks, marked by irregular geometries and narrow streets. These visual distinctions offer spatial context for understanding how urban morphology may influence traffic dynamics, routing behavior, and vehicle emissions.

To further leverage the spatial richness of these satellite images, we extracted numerical representations through a computer vision pipeline using a pretrained ResNet-18 convolutional neural network from the \texttt{torchvision} package~\cite{paszke2019pytorch} to generate 512-dimensional feature vectors from each town.

\begin{figure}[htbp]
    \centering
    \includegraphics[width=0.7\textwidth]{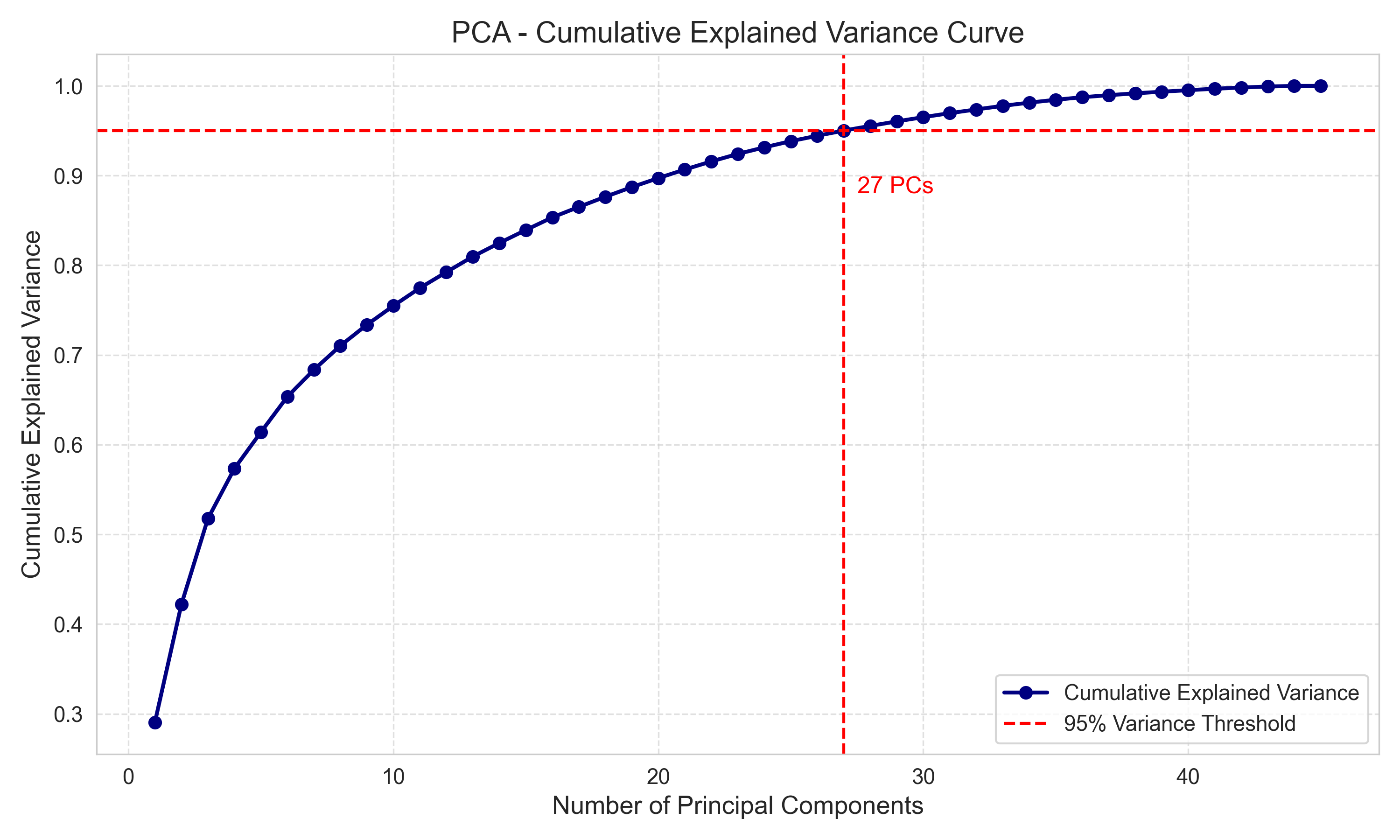}
    \caption{Cumulative explained variance curve from PCA applied to ResNet-encoded image features. The red dashed line indicates the 95\% variance threshold, with the corresponding number of components labeled.}
    \label{fig:pca_curve}
\end{figure}

We applied PCA ~\cite{jolliffe2002principal} to reduce feature dimensionality while retaining 95\% of the cumulative variance. The resulting compact vectors encode town layouts as structured inputs for downstream modeling. Figure~\ref{fig:pca_curve} shows the variance curve and component selection.

To ensure data quality and suitability for subsequent analysis, raw trajectory data from public GPS traces is subjected to a rigorous preprocessing pipeline. This process primarily aims to achieve a uniform temporal resolution and effectively handle missing observations. A consistent temporal resolution of one minute is established for all trajectories. Furthermore, extended discontinuities within a single trajectory are systematically excluded from the dataset. For shorter data gaps, specifically those not exceeding 180 seconds, a local regression technique, as detailed by ~\cite{toledo2007estimation}, is employed for imputation.

\begin{figure}[htbp]
    \centering
    \includegraphics[width=\linewidth]{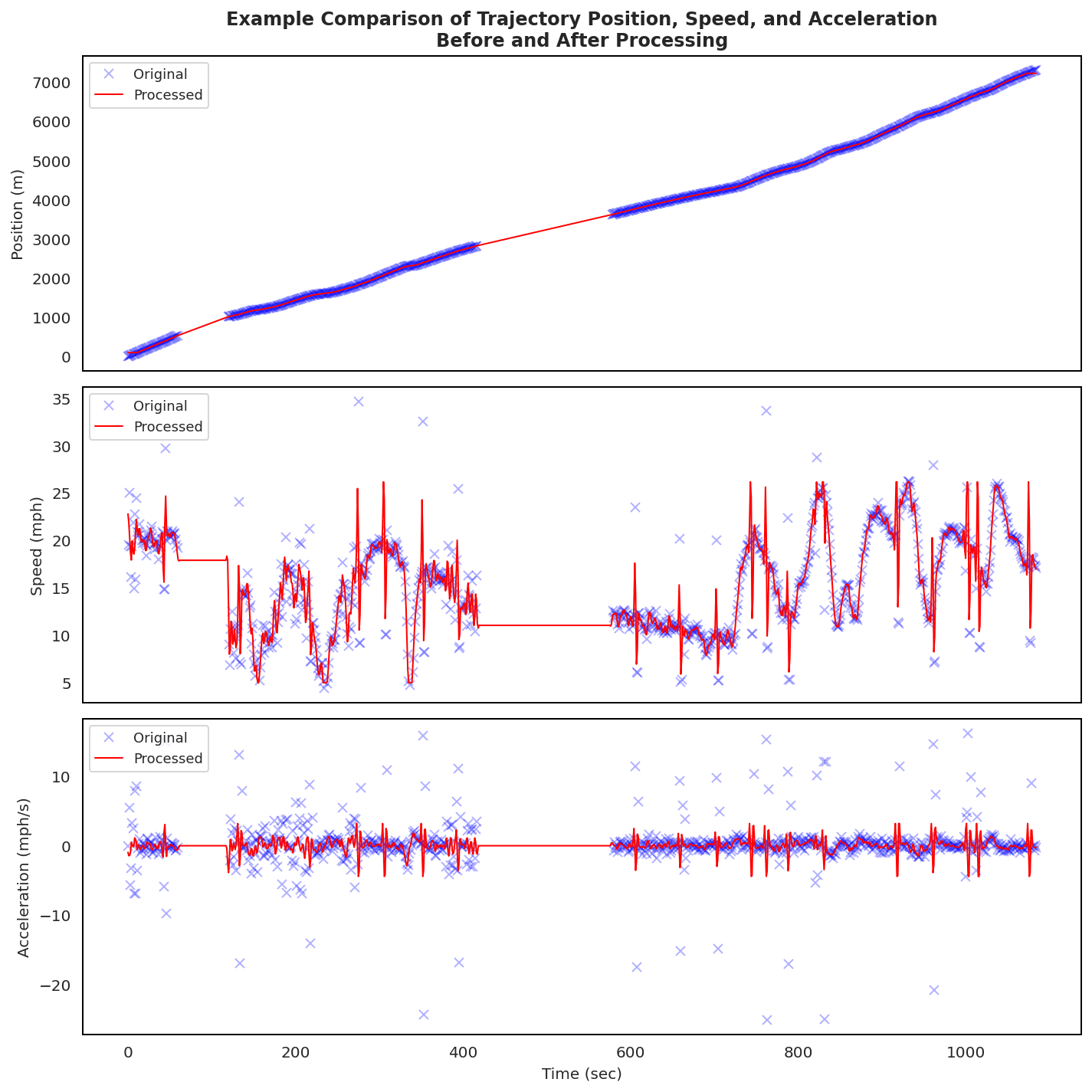}
    \caption{A sample vehicle trajectory data before and after processing.}
    \label{fig:traj_compare}
\end{figure}

The necessity raw trajectory data processing is evident in Figure~\ref{fig:traj_compare}, which shows a sample vehicle trajectory both in its raw and processed states. The unprocessed trajectory, in blue line, contains considerable temporal gaps. Moreover, the high temporal resolution, combined with these inconsistencies, leads to noisy and unreliable derivations of speed and acceleration. Following the processing steps, including gap imputation, the resulting speed and acceleration profiles, shown in red, become smoother, enabling more accurate and stable downstream tasks.


\subsection{Training}

Available trajectory data from multiple peak-period bi-hourly time windows, such as 7–9 AM, 9–11 AM, 4–6 PM, 6–8 PM, and 8–10 PM was used. To ensure data quality and reliability, only road segments with a physical length greater than 50 meters and cumulative trajectory durations exceeding 120 seconds were included in the training process. After preprocessing and filtering, the dataset consisted of approximately 6000 samples.

The dataset, consisting of both numerical and categorical features, was divided into training and testing subsets using an 80:20 ratio. The target variable included 23 output components representing operating mode distributions. Numerical features such as peak hour flow, road length, number of lanes, and actual speed were standardized using the \texttt{StandardScaler} from the \texttt{sklearn} library~\cite{scikit-learn}, which transforms features to have zero mean and unit variance~\cite{standardization}. Categorical variables (e.g., directionality, road type, and urban type) were one-hot encoded, and the resulting numerical and categorical features were concatenated. For links with missing one-way information, the directionality of corresponding OSM edges is used to infer whether the link is one-way or two-way. For links lacking speed limit data, the missing values are imputed using the average speed limit of roads with the same classification.

The training and testing data were converted into \texttt{TensorDataset} objects and fed into \texttt{DataLoader} with a batch size of 32~\cite{torch_dataloader}, allowing the model to process the data in mini-batches and shuffle the training data during each epoch to improve generalization. The input layer size is set to total number of processed features, and mean square error (MSE) is used as a loss function. The Adam optimizer~\cite{adam} with a learning rate of 0.001 is employed for parameter updates. The model is trained for 1000 epochs. A dropout ratio of 0.3, in both shared feature extraction and module’s network layers, is employed during the training process. During each epoch, both training and testing losses were recorded.

\begin{figure}[htbp]
    \centering
    \includegraphics[width=5in]{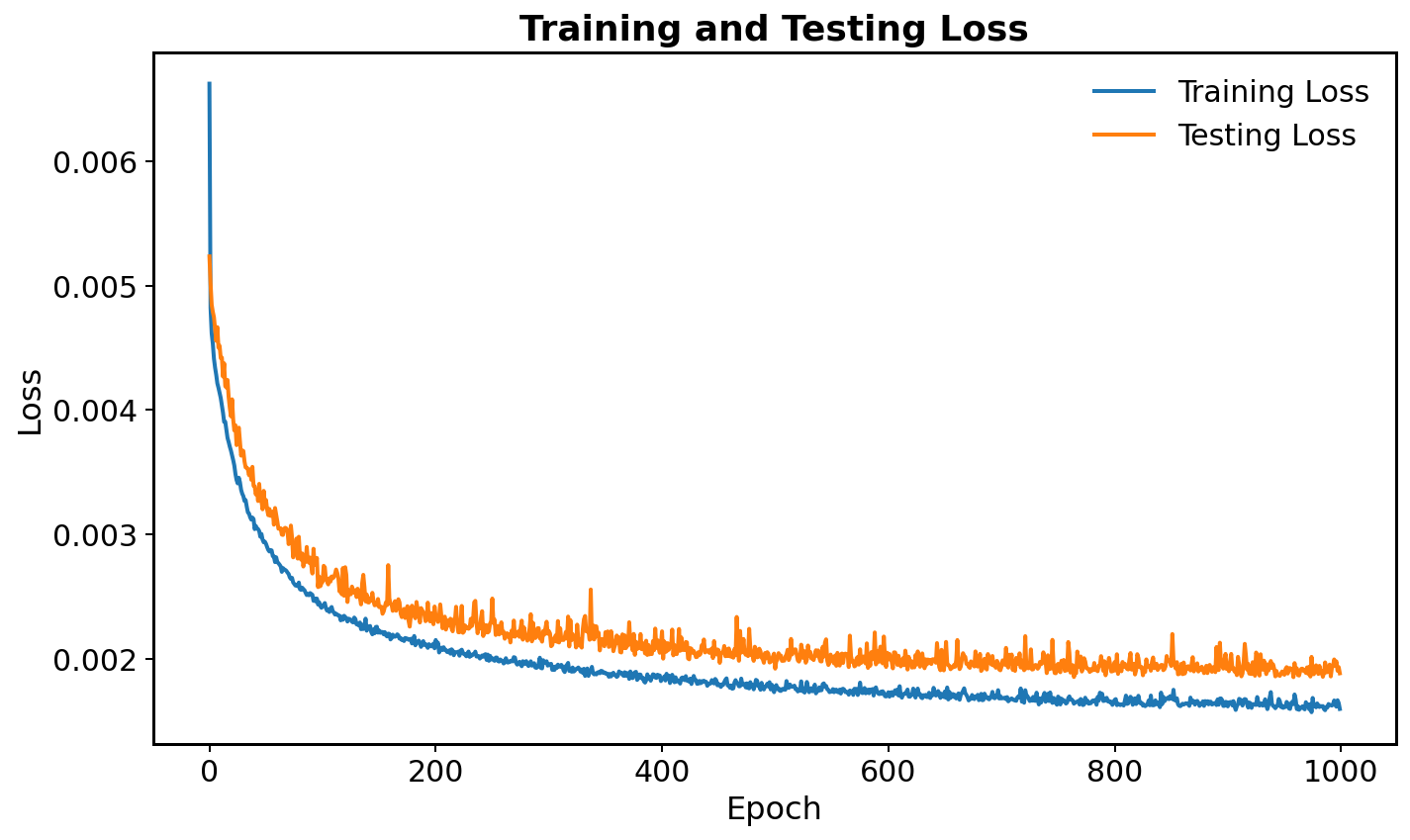}
    \caption{Training and testing loss of the proposed model.}
    \label{fig:loss_curve}
\end{figure}

As shown in Figure~\ref{fig:loss_curve}, the training and testing losses declined rapidly in the first 200 epochs, suggesting effective learning during the early stages. Afterward, the training loss continued to decrease gradually.

Model training was conducted on a system equipped with an NVIDIA GeForce RTX 4060 Ti GPU.

\subsection{Evaluation}
This study evaluates the performance of the proposed model by comparing emissions estimated using three different approaches. First, the detailed vehicle trajectories from the publicly available OSM-based datasets are processed into second-by-second records with instantaneous speed and acceleration. The operating mode distributions are then directly calculated from these trajectories and used in MOVES to generate accurate emission estimates, serving as the ground truth. 

Second, the default MOVES methodology is applied. In this approach, average link speeds are fed into MOVES, which internally selects appropriate driving cycles and derives corresponding operating mode distributions. These default distributions are subsequently used for emission estimation.

Third, the proposed model is used. The trained Modular Neural Network (MNN) takes link-level features as input and predicts the operating mode distribution for each road segment, which is then used with MOVES for emission estimation.

To evaluate model performance, we use the root mean square error (RMSE) and coefficient of determination ($R^2$) from Python’s \texttt{sklearn} library to compare the results from the proposed MNN model and the MOVES default method against the ground truth emissions. The results show that the proposed model generally outperforms the default MOVES procedure, particularly in operating modes with higher driving condition fractions.

\begin{figure}[htbp]
    \centering
    \begin{subfigure}[b]{0.48\linewidth}
        \centering
        \includegraphics[width=\textwidth]{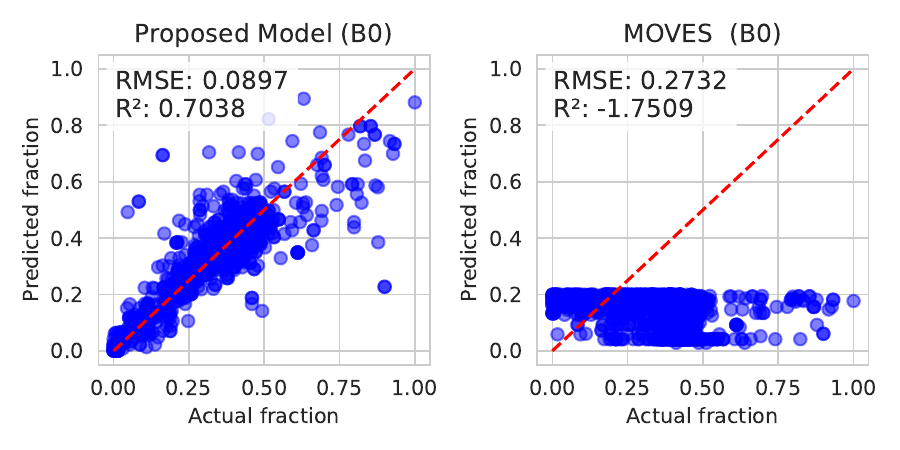}
        \caption{\fontsize{8}{12}\selectfont Estimated vs true operating mode fractions of Bin 0}
        \label{fig:7(a)}
    \end{subfigure}
    \hfill
    \begin{subfigure}[b]{0.48\textwidth}
        \centering
        \includegraphics[width=\textwidth]{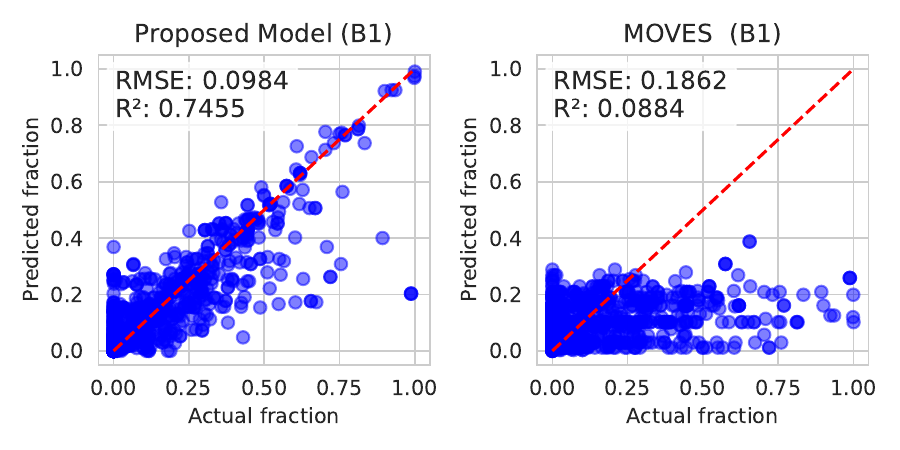}
        \caption{\fontsize{8}{12}\selectfont Estimated vs true operating mode fractions of Bin 1}
        \label{fig:7(b)}
    \end{subfigure}
    \\
    \begin{subfigure}[b]{0.48\textwidth}
        \centering
        \includegraphics[width=\textwidth]{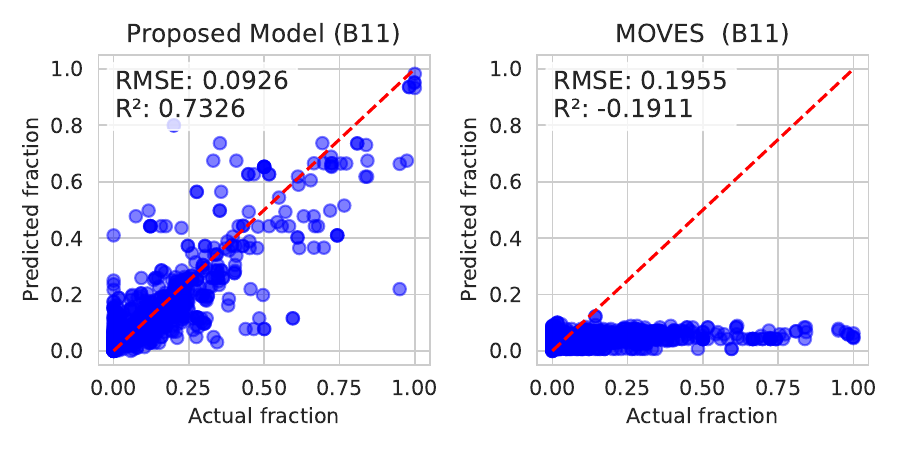}
        \caption{\fontsize{8}{12}\selectfont Estimated vs true operating mode fractions of Bin 11}
        \label{fig:7(c)}
    \end{subfigure}
    \hfill
    \begin{subfigure}[b]{0.48\textwidth}
        \centering
        \includegraphics[width=\textwidth]{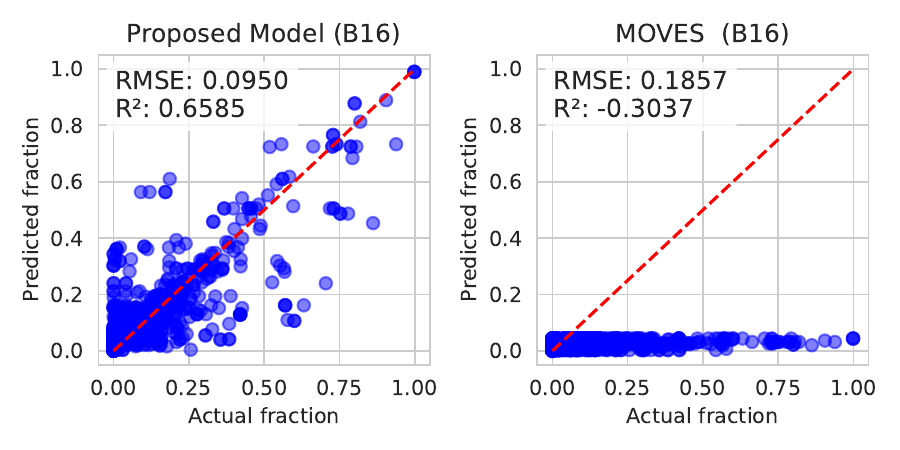}
        \caption{\fontsize{8}{12}\selectfont Estimated vs true operating mode fractions of Bin 16}
        \label{fig:7(d)}
    \end{subfigure}
    \\
    \begin{subfigure}[b]{0.48\textwidth}
        \centering
        \includegraphics[width=\textwidth]{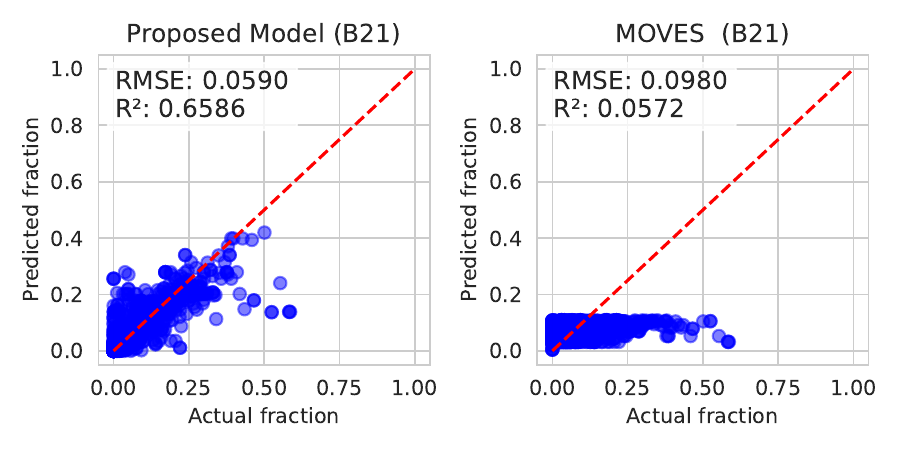}
        \caption{\fontsize{8}{12}\selectfont Estimated vs true operating mode fractions of Bin 21}
        \label{fig:7(e)}
    \end{subfigure}
    \hfill
    \begin{subfigure}[b]{0.48\textwidth}
        \centering
        \includegraphics[width=\textwidth]{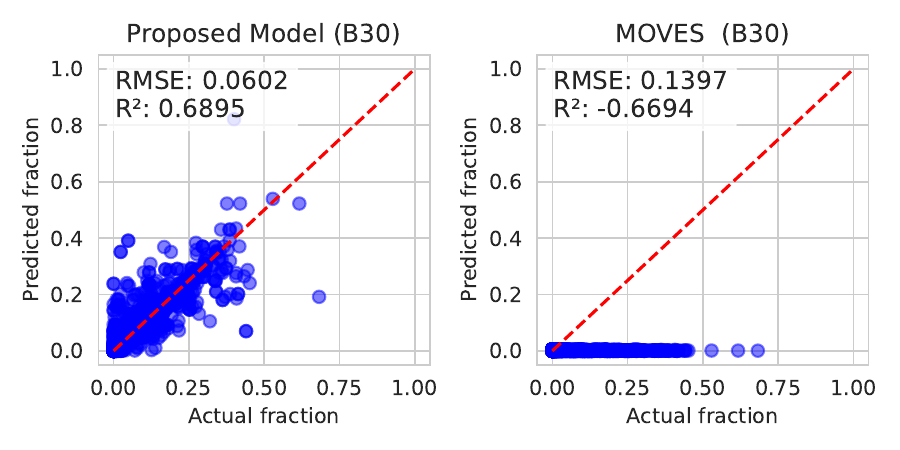}
        \caption{\fontsize{8}{12}\selectfont Estimated vs true operating mode fractions of Bin 30}
        \label{fig:7(f)}
    \end{subfigure}
    \\
    \begin{subfigure}[b]{0.48\textwidth}
        \centering
        \includegraphics[width=\textwidth]{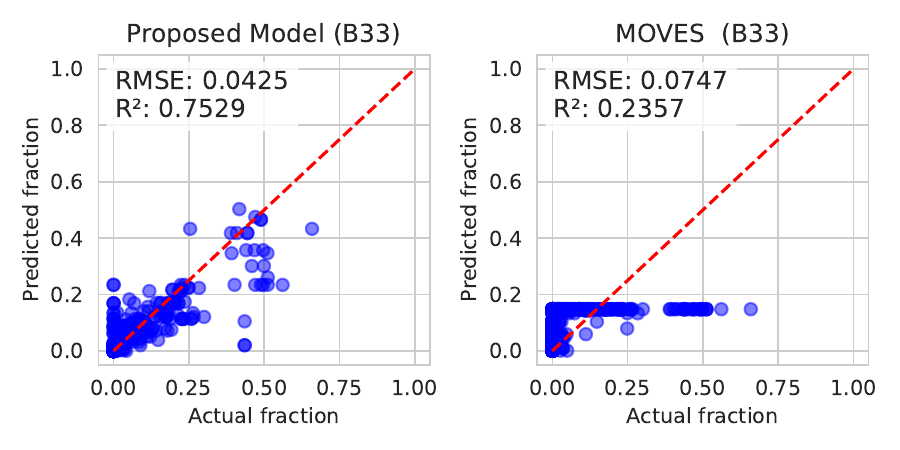}
        \caption{\fontsize{8}{12}\selectfont Estimated vs true operating mode fractions of Bin 33}
        \label{fig:7(g)}
    \end{subfigure}
    \hfill
    \begin{subfigure}[b]{0.48\textwidth}
        \centering
        \includegraphics[width=\textwidth]{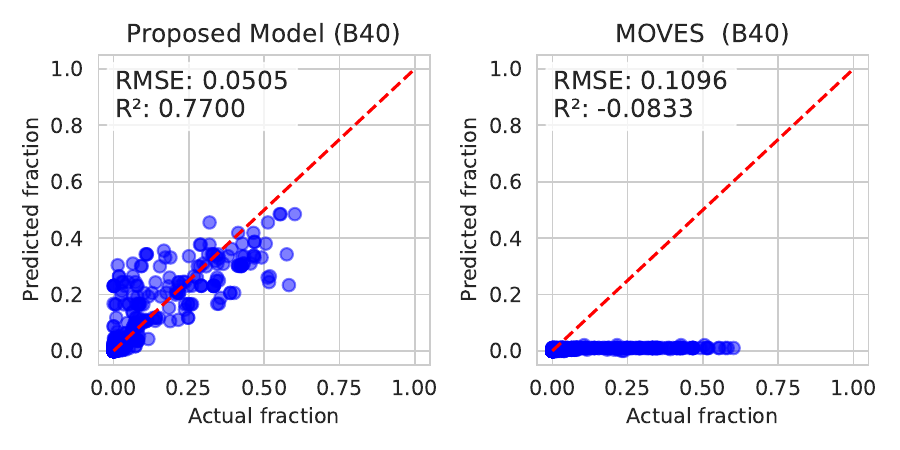}
        \caption{\fontsize{8}{12}\selectfont Estimated vs true operating mode fractions of Bin 40}
        \label{fig:7(h)}
    \end{subfigure}
    \caption{Comparison of operating model fractions obtained using the proposed model and MOVES against ground truth.}
    \label{fig:figure7}
\end{figure}

Figure~\ref{fig:figure7} compares the estimated operating mode fractions from the proposed model and MOVES against the ground truth. Eight operating mode bins with nonzero fractions in the ground truth dataset are selected to evaluate the model's predictive ability. Figures~\ref{fig:7(a)} and~\ref{fig:7(b)} show the comparison for Bins 0 and 1 (braking/idling module), Figures~\ref{fig:7(c)} and~\ref{fig:7(d)} for Bins 11 and 16 (low-speed module), Figures~\ref{fig:7(e)} and~\ref{fig:7(f)} for Bins 21 and 30 (moderate-speed module), and Figures~\ref{fig:7(g)} and~\ref{fig:7(h)} for Bins 33 and 40 (high-speed module).

The results indicate that for operating modes with generally nonzero fractions, the proposed model achieves strong performance, with predicted values closely aligning with actual data. Most $R^2$ scores are around 0.7, and RMSE values remain relatively consistent and low. In contrast, the default MOVES estimation produces very low $R^2$ values, often close to zero, and for bins 0, 11, 16, 30, and 40, the $R^2$ values are even negative. This demonstrates that the model improves the accuracy of operating mode distribution estimation. Despite the variation in actual fractions across bins, the proposed model maintains a balanced level of prediction accuracy, suggesting that it can reliably estimate the distribution of various operating modes, as long as the operating modes are sufficiently represented in the dataset.

\begin{figure}[htbp]
    \centering
    \includegraphics[width=\textwidth]{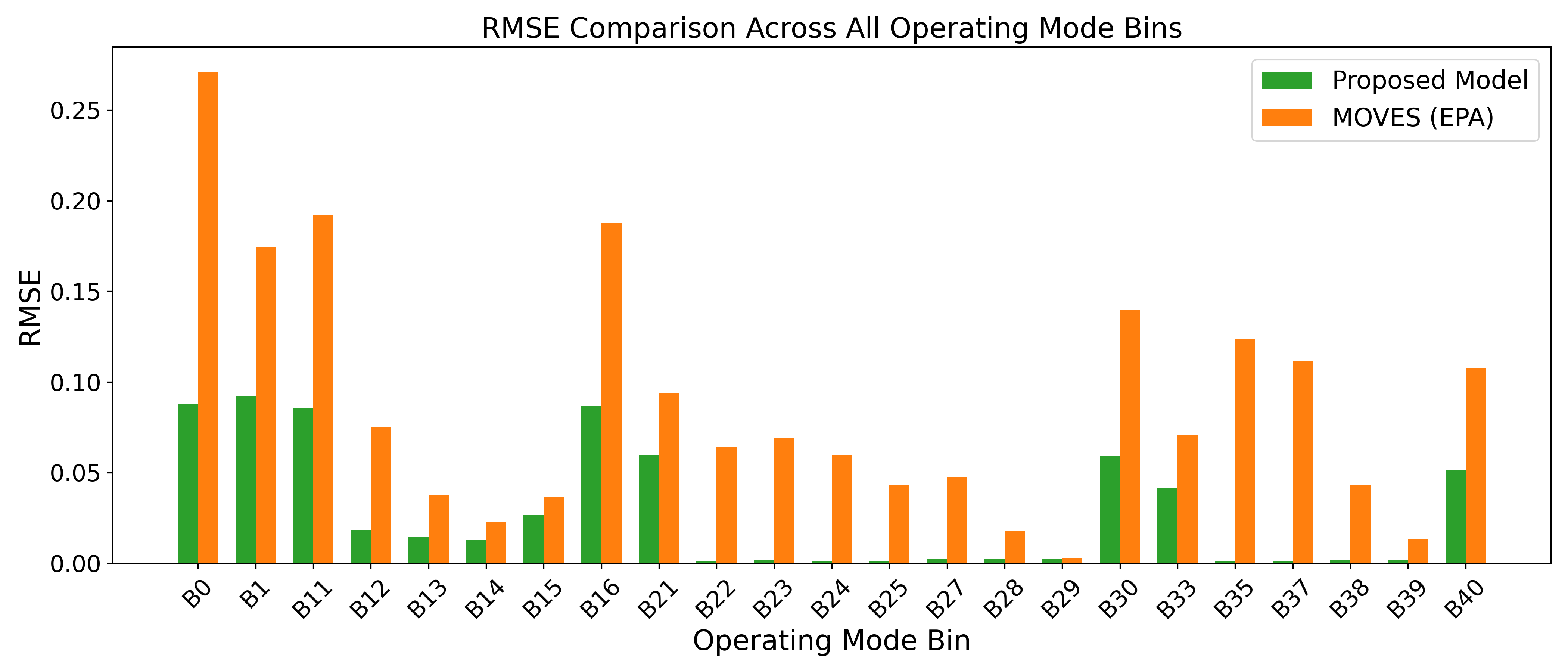}
    \caption{Comparison of RMSE across all operating mode bins between the proposed model and the EPA MOVES default method. The green bars represent RMSE from the proposed model, while the orange bars represent RMSE from MOVES. Lower values indicate better agreement with ground truth.}
    \label{fig:rmse_bar_all}
\end{figure}

Figure~\ref{fig:rmse_bar_all} presents a comparison of root mean square error (RMSE) values for each operating mode bin, contrasting the performance of the proposed model with the default operating mode distribution method in EPA’s MOVES. As shown in the figure, the proposed model consistently achieves lower RMSE across all operating mode bins, indicating improved alignment with ground truth distributions. The performance advantage is particularly pronounced for bins 22 through 29 and bins 35 through 39, where the true fractions are extremely low—often close to zero across most road segments. Despite this sparsity, the proposed model yields near-zero RMSE in these bins, demonstrating its ability to accurately predict rare operating modes. In contrast, the MOVES default method produces substantially higher errors in these bins.

\begin{figure}[htbp]
    \centering
    \begin{subfigure}[b]{0.48\linewidth}
        \centering
        \includegraphics[width=\textwidth]{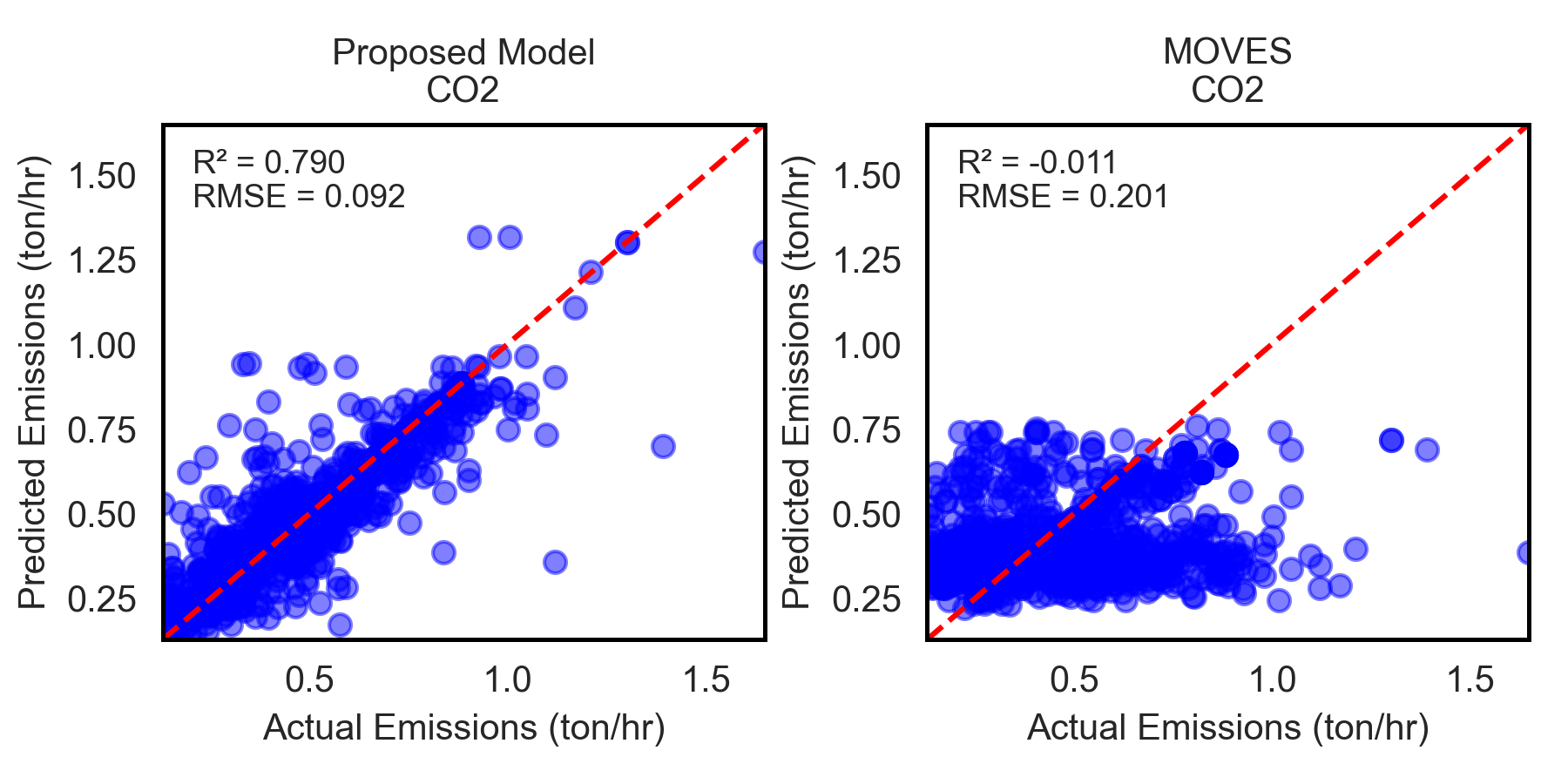}
        \caption{\fontsize{8}{12}\selectfont Estimated vs true CO\textsubscript{2} emission}
        \label{fig:8(a)}
    \end{subfigure}
    \hfill
    \begin{subfigure}[b]{0.48\textwidth}
        \centering
        \includegraphics[width=\textwidth]{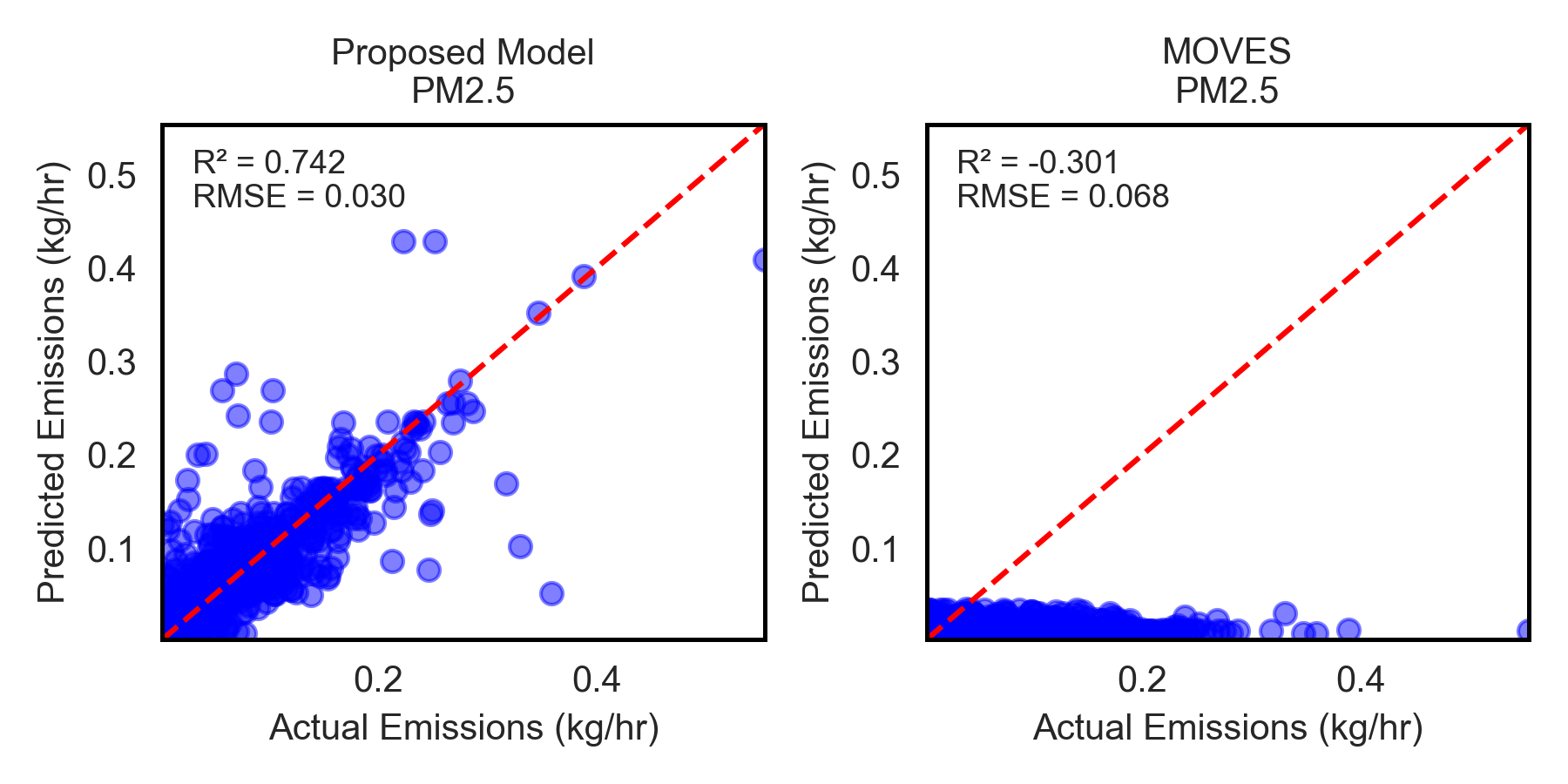}
        \caption{\fontsize{8}{12}\selectfont Estimated vs true PM\textsubscript{2.5} emission}
        \label{fig:8(b)}
    \end{subfigure}

    \begin{subfigure}[b]{0.48\textwidth}
        \centering
        \includegraphics[width=\textwidth]{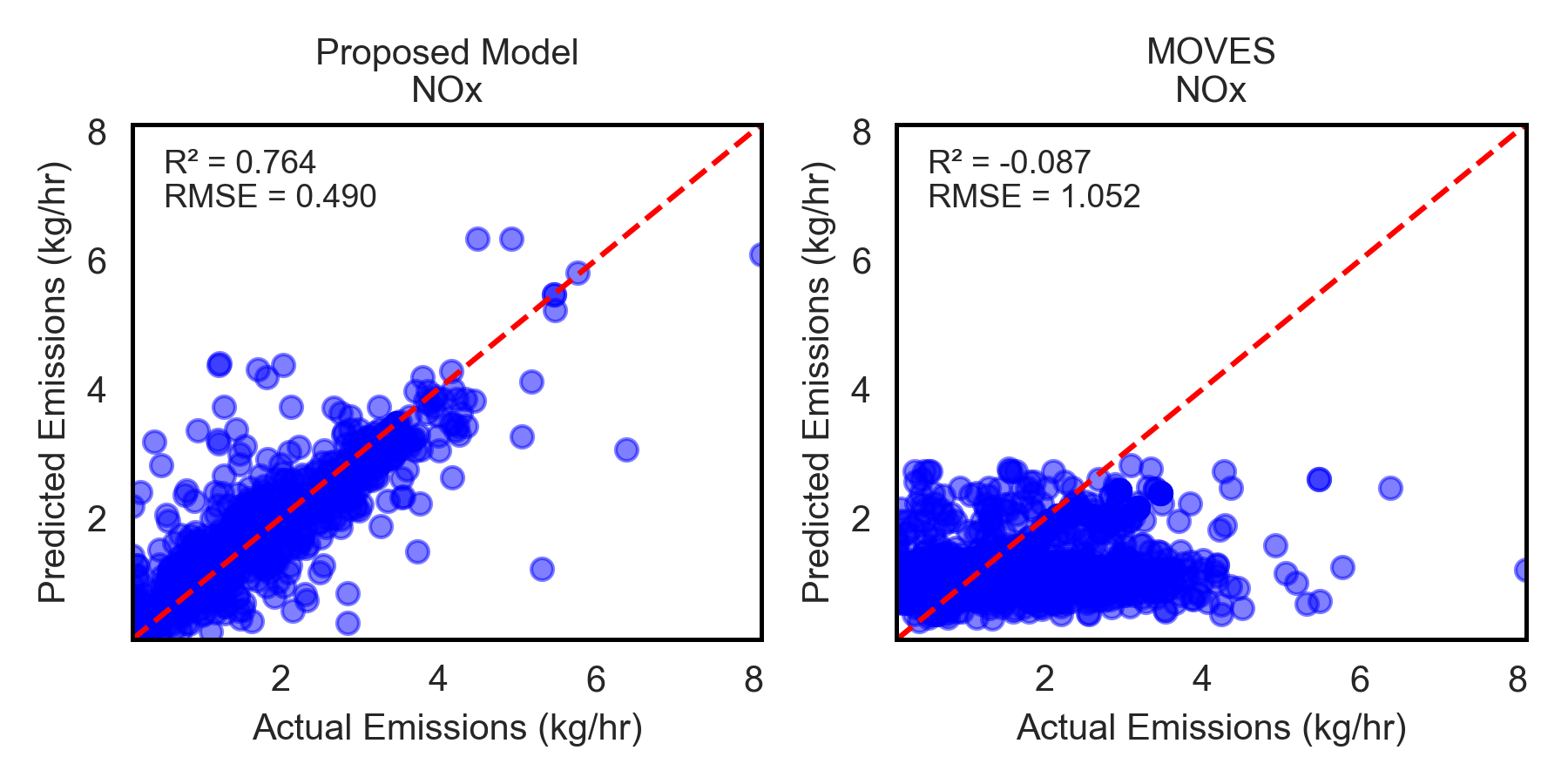}
        \caption{\fontsize{8}{12}\selectfont Estimated vs true NO\textsubscript{x} emission}
        \label{fig:8(c)}
    \end{subfigure}
    \hfill
    \begin{subfigure}[b]{0.48\textwidth}
        \centering
        \includegraphics[width=\textwidth]{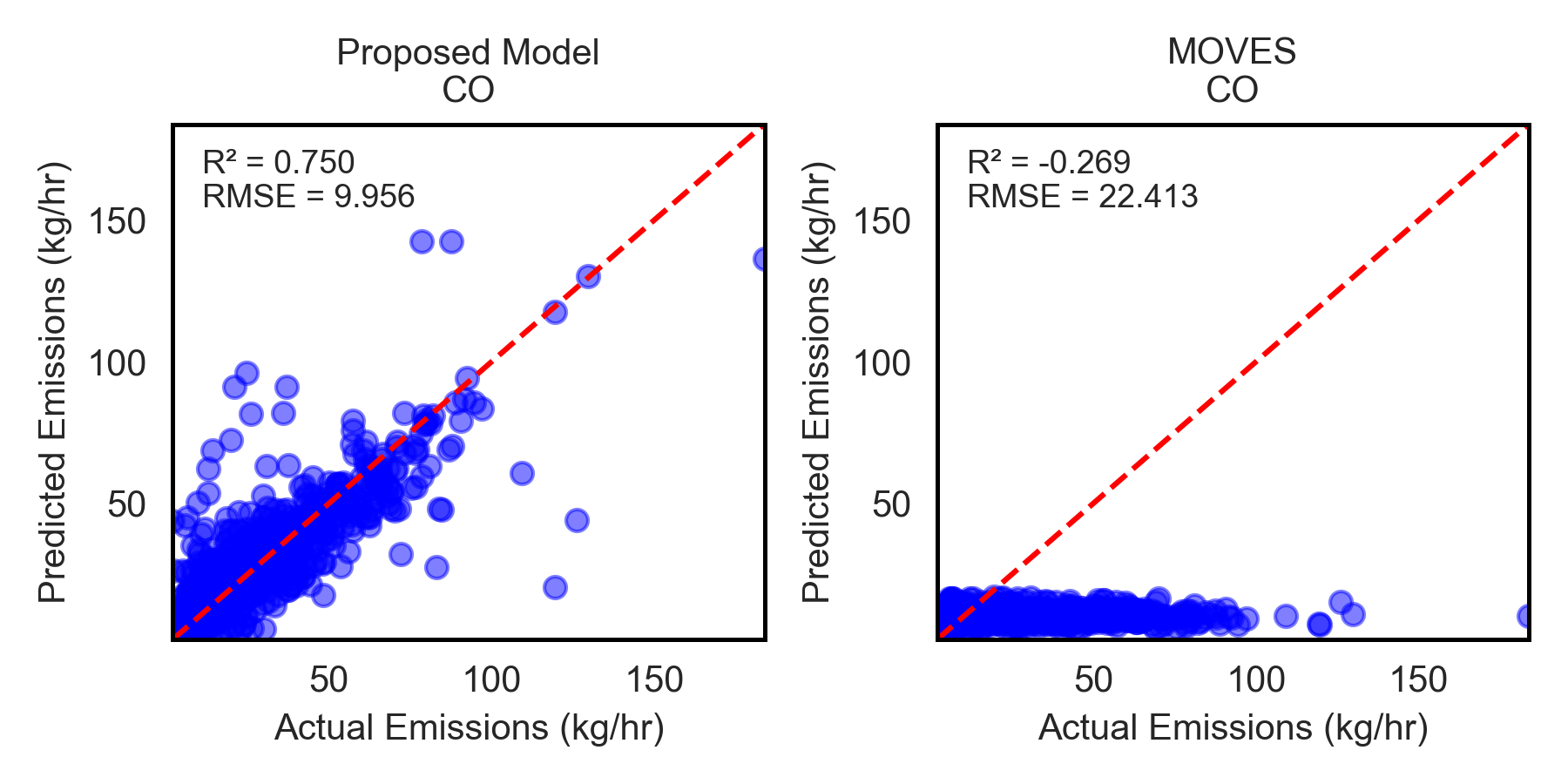}
        \caption{\fontsize{8}{12}\selectfont Estimated vs true CO emission}
        \label{fig:8(d)}
    \end{subfigure}

    \caption{Comparison of estimated traffic emissions using the proposed model and MOVES against ground truth.}
    \label{fig:figure8}
\end{figure}

We employ three distinct operating mode distributions - (1) ground truth derived from trajectory data, (2) MOVES default driving cycle outputs, and (3) proposel model's predictions - as inputs to the MOVES framework to estimate emissions for key pollutants. Figure~\ref{fig:figure8} presents a comparison between the proposed model and MOVES default approach in estimating hourly emissions of four key pollutants: CO$_2$, PM$_{2.5}$, SO$_2$, CH$_4$, NO$_x$, and CO against the ground truth. Each subplot compares estimated emissions with the true emissions (derived from MOVES using ground truth operating mode distributions) across 1069 test links, with the red dashed line indicating the 1:1 reference line.

The proposed model consistently improves the default MOVES model estimates across all four pollutants, as evidenced by lower root mean square error (RMSE) values and higher $R^2$ scores in each subplot. For instance, in estimating CO$_2$ emissions (Figure~\ref{fig:figure8}(a)), the proposed model achieves an $R^2$ of 0.79 compared to -0.011 for MOVES, with a significantly lower RMSE. Similar improvements are observed for PM$_{2.5}$, NO$_x$, and CO, as shown in Figures~\ref{fig:figure8}(b)-(f). MOVES default estimates show almost no correlation with ground truth, while the proposed model demonstrates strong alignment.

\begin{figure}[htbp]
    \centering
    \includegraphics[width=0.6\textwidth]{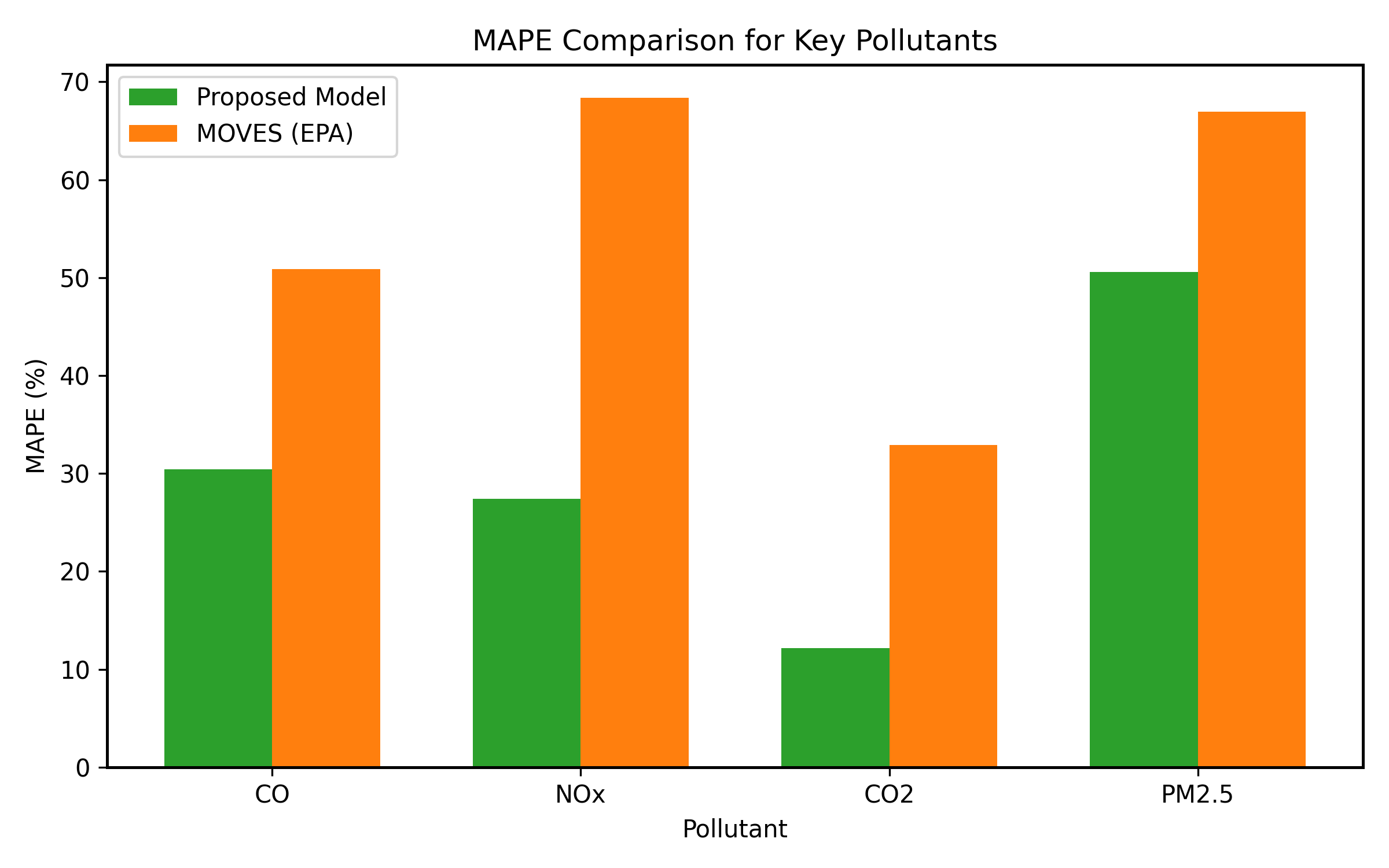}
    \caption{Comparison of Mean Absolute Percentage Error (MAPE) between the proposed model and MOVES (EPA) for key pollutants including CO, NO\textsubscript{x}, CO\textsubscript{2}, and PM\textsubscript{2.5}.}
    \label{fig:MAPE_percent_comparison}
\end{figure}

Based on the comparative performance of the proposed model and the MOVES default method, Figure~\ref{fig:MAPE_percent_comparison} illustrates the mean absolute percentage error (MAPE) for four key pollutants: CO, NO$_x$, CO$_2$, and PM$_{2.5}$. The proposed model consistently achieves substantially lower MAPE across all pollutants. Specifically, MAPE is reduced by 40.20\% for CO (from 50.90\% to 30.44\%), 59.86\% for NO$_x$ (from 68.34\% to 27.43\%), 63.03\% for CO$_2$ (from 32.91\% to 12.17\%), and 24.41\% for PM$_{2.5}$ (from 66.95\% to 50.61\%). In addition to MAPE, the RMSE reduction compared to the MOVES default approach is 55.58\% for CO, 53.39\% for NO$_x$, 54.44\% for CO$_2$, and 55.50\% for PM$_{2.5}$. 


The results confirm that improved estimation of traffic emission estimates is possible using open source data. The proposed modular neural network model offers a scalable and data-efficient alternative for large-scale traffic emission modeling only using readily available open source data.

\section{Conclusion}

This study presents a neural network-based framework to enhance the MOVES estimation of link-level emissions across citywide traffic networks, leveraging readily available open-source data. The traditional MOVES project-level approach, which relies on either actual trajectory data or default driving cycles to determine operating mode distributions, is often data-intensive and hinders scalability for network-wide applications. The proposed model addresses this limitation by replacing the operating mode distribution estimation step with a neural network trained on publicly available vehicle trajectories. Once trained, the model requires only easily accessible data to quickly and accurately estimate operating modes, allowing for efficient, large-scale emissions calculations without compromising accuracy. 



The result shows that the proposed model demonstrates improved performance over the default MOVES approach in estimating operating mode distributions. RMSE values are consistently low, and $R^2$ scores are mostly around 0.7 for commonly observed bins, while MOVES’ default estimates yield near-zero or negative $R^2$ values in many bins, including Bin 0, 11, 16, 30, and 40. Emission estimations based on the model's predicted distributions align much more closely with ground truth, achieving an $R^2$ of 0.79 for CO$_2$ emissions compared to -0.011 from MOVES. Similar improvements are observed across pollutants including PM$_{2.5}$, NO$_x$, and CO. The proposed framework outperforms for low activity bins i.e., bins with lower fractions, particularly for bins  22 through 29 and bins 35 through 39. In contrast, the EPA MOVES default approach tends to overestimate emissions for these low-activity bins. Moreover, when aggregated across the entire test dataset, the proposed model achieves more than a 50\% reduction in RMSE for emissions compared to the MOVES default approach across all four key pollutants.

Building on the model's promising performance in the Boston Metropolitan area, future research will focus on its application to a wider range of urban environments. A primary direction is to validate and refine the model's performance in cities with different traffic patterns, network structures, and vehicle fleets. This will involve developing a more generalizable framework that can be easily adapted to various regional contexts. Another avenue for future work is to integrate computer vision pipelines designed for satellite imagery, to derive better representations of traffic flow characteristics and network topology.

\section{Acknowledgements}
This work was supported by the Northeastern University iSUPER Impact Engine. The authors are grateful for the support of Northeastern University. Any opinions, findings, conclusions, or recommendations expressed in the paper are those of the authors and do not necessarily reflect the views of the funding agencies. The Large Language Model ChatGPT 3.5/4o was used to improve grammar, phrasing, and clarity of the manuscript.

\section{AUTHOR CONTRIBUTIONS}
The authors confirm contribution to the paper as follows: study conception and design: L. Wang, M. Usama, H. N. Koutsopoulos, Z. He; data collection: L. Wang, M. Usama; analysis and interpretation of results: L. Wang, M. Usama, H. N. Koutsopoulos and Z. He; draft manuscript preparation: L. Wang, M. Usama and H. N. Koutsopoulos. All authors reviewed the results and approved the final version of the manuscript.

\section{Conflict of Interest Statement}
Zhengbing He is a member of Transportation Research Record's Editorial Board. Apart from this, the authors declared no potential conflicts of interest with respect to the research, authorship, and/or publication of this article.

\newpage

\bibliographystyle{trb}
\bibliography{trb_template}
\end{document}